\documentclass[11pt, a4paper, goog]{google}

\usepackage[authoryear, sort&compress, round]{natbib}
\usepackage{graphicx}
\usepackage{booktabs}
\usepackage{multirow}
\usepackage{tabularx}
\usepackage{hyperref}
\usepackage{xspace}
\usepackage{cleveref}
\usepackage{subcaption}

\usepackage{xcolor}

\usepackage{amsmath}
\DeclareMathOperator*{\argmax}{argmax}
\DeclareMathOperator*{\argmin}{argmin}

\usepackage{makecell}
\usepackage[table]{xcolor} 

\definecolor{best_color}{HTML}{90EE90}  
\definecolor{second_color}{HTML}{FFFF99}

\definecolor{mycolor_blue}{RGB}{210, 230, 255}
\definecolor{mycolor_yellow}{RGB}{255, 255, 200}
\definecolor{mycolor_red}{RGB}{255, 200, 200}

\newcommand{\vqqa}{%
  \textbf{%
    \textcolor{red!70!white}{V}%
    \textcolor{green!70!blue}{Q}%
    \textcolor{yellow!70!purple}{Q}
    \textcolor{blue!80!black}{A}%
  }\xspace%
}

\newcommand{\vqqafull}{%
  \textcolor{red!70!white}{\textbf{V}}ideo\xspace%
  \textcolor{green!70!blue}{\textbf{Q}}uality\xspace%
  \textcolor{yellow!70!purple}{\textbf{Q}}uestion\xspace%
  \textcolor{blue!80!black}{\textbf{A}}nswering%
  \xspace
}

\usepackage[most]{tcolorbox}
\usepackage{listings}
\usepackage{xcolor}

\lstdefinestyle{markdown}{
  basicstyle=\ttfamily\small,
  breaklines=true,
  columns=fullflexible,
  keepspaces=true
}

\newtcolorbox{agentbox}[2][]{
  enhanced,
  breakable,                     
  colback=white,
  colframe=black!70,
  boxrule=0.7pt,
  arc=2pt,
  left=8pt,right=8pt,top=8pt,bottom=8pt,
  fonttitle=\bfseries,
  colbacktitle=gray!15,
  coltitle=black,
  title={#2},
  attach boxed title to top left={yshift=-2mm, xshift=2mm},
  boxed title style={
      boxrule=0.6pt,
      arc=2pt,
      colframe=black!70,
      colback=gray!15
  },
  #1
}

\uselogo{} 

\title{\textls[-25]{\vqqa: An Agentic Approach for Video~Evaluation} and Quality Improvement}

\correspondingauthor{yiwensong@google.com, yalesong@google.com}

\reportnumber{} 

\renewcommand{\today}{}

\author[1]{Yiwen Song}
\author[1]{Tomas Pfister}
\author[1]{Yale Song}

\affil[1]{\thepa{}{}}

\begin{abstract}
  Despite rapid advancements in video generation models, aligning their outputs with complex user intent remains challenging. Existing test-time optimization methods are typically either computationally expensive or require white-box access to model internals. To address this, we present \vqqa (\vqqafull), a unified, multi-agent framework generalizable across diverse input modalities and video generation tasks. By dynamically generating visual questions and using the resulting Vision-Language Model (VLM) critiques as semantic gradients, \vqqa replaces traditional, passive evaluation metrics with human-interpretable, actionable feedback. This enables a highly efficient, closed-loop prompt optimization process via a black-box natural language interface. Extensive experiments demonstrate that \vqqa effectively isolates and resolves visual artifacts, substantially improving generation quality in just a few refinement steps. Applicable to both text-to-video (T2V) and image-to-video (I2V) tasks, our method achieves absolute improvements of \textbf{+11.57\%} on T2V-CompBench and \textbf{+8.43\%} on VBench2 over vanilla generation, significantly outperforming state-of-the-art stochastic search and prompt optimization techniques.\\ \\Project Page: \url{https://yiwen-song.github.io/vqqa/}
\end{abstract}

\begin{document}

\maketitle

\section{Introduction}
\label{sec:intro}

Driven by breakthroughs in diffusion and transformer architectures, visual generative models have revolutionized dynamic, high-resolution scene rendering~\citep{peebles2023scalable, blattmann2023align, openai2024sora, deepmind2025veo3, wan2025, yang2024cogvideox, HaCohen2024LTXVideo, team2025kling}. However, aligning these models with complex human intent remains challenging~\citep{ma2025controllable}. Users frequently encounter compositional errors, temporal inconsistencies, and physical hallucinations~\citep{liu2023fetv}, which require tedious, trial-and-error prompt engineering.

Furthermore, evaluation methods have not kept pace with model development. Early metrics~\citep{unterthiner2018towards, salimans2016improved} measure basic visual distributions but miss complex compositional alignment. Comprehensive benchmarks~\citep{liu2023evalcrafter, liao2024evaluation, meng2024towards, ling2025vmbench} address this but require large, specialized model ensembles~\citep{huang2024vbench, sun2025t2v, huang2025vbench++, zheng2025vbench2}, creating significant computational overhead. Alternatively, regressors or Vision-Language Models (VLMs) can predict human opinion scores~\citep{kou2024subjective, ge2025lmm, lin2024evaluating}. Yet, these systems act as passive observers, lacking the flexibility to adapt to new tasks or provide actionable feedback to correct generations.

Concurrently, agentic workflows, where AI systems autonomously plan, execute, and refine their own output, show promise in overcoming traditional generation limits~\citep{genartist2024, mccd_2025_cvpr, wu2025automated}. However, existing video test-time optimization typically relies on computationally intensive selection (e.g., VISTA's pairwise tournaments~\citep{long2025vista}) or requires white-box access to model internals~\citep{dalal2025one}. There is a critical need for an interpretable, closed-loop system that diagnoses visual flaws and iteratively refines videos via a black-box natural language interface.

To address this, we propose \textbf{\vqqa} (\vqqafull), a unified multi-agent framework for holistic evaluation and iterative prompt refinement. By employing a dynamic question-answering paradigm instead of static rubrics, \vqqa adapts seamlessly to diverse conditional tasks (e.g., T2V, I2V), transforming evaluation from a passive metric into actionable feedback.

\vqqa operates via three specialized agents: \textit{Question Generation} formulates targeted visual queries from inputs; \textit{Question Answering} evaluates the video to isolate critical flaws; and \textit{Prompt Refinement} uses these diagnostics as a \textit{semantic gradient}~\citep{yuksekgonul2024textgrad,lee2025feedback} to optimize the prompt for the next iteration.

To prevent \textit{semantic drift} during refinement, \vqqa employs a Global Selection mechanism where a VLM evaluates candidates against the initial prompt. Coupled with a dynamic stopping criterion, this maximizes visual quality and minimizes compute overhead. Furthermore, \vqqa is strictly model-agnostic, generalizing across models and modalities without task-specific fine-tuning.

\begin{figure*}[t]
    \centering
    \includegraphics[width=\linewidth]{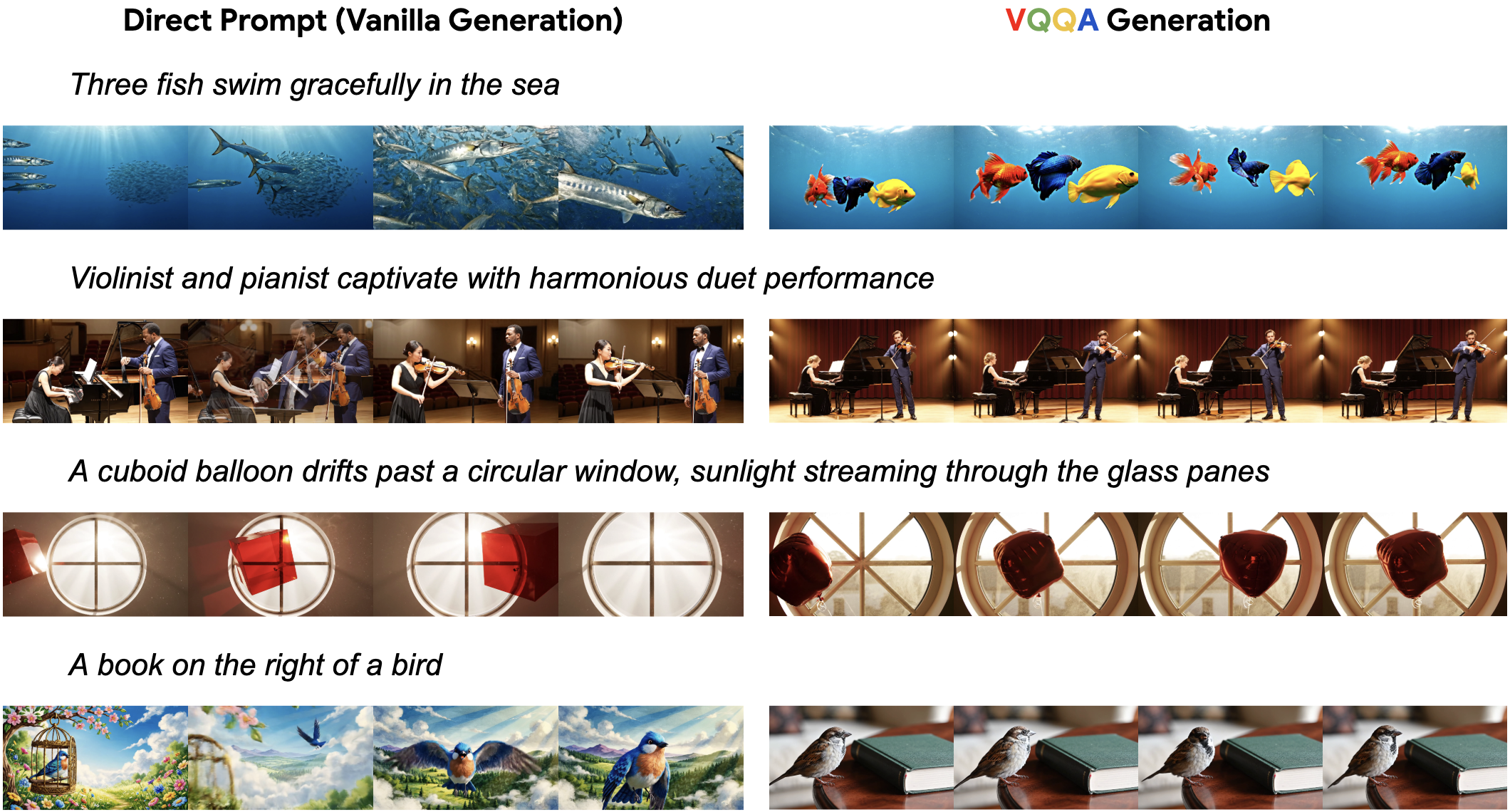}
    \caption{\textbf{Videos generated by \vqqa}, showing improvements in categories such as numeracy, interaction, dynamic attributes, and spatial relationship.}
    \label{fig:vqqa_main_examples}
\end{figure*}

Our main contributions are:
\begin{itemize}
    \item We propose \vqqa, a multi-agent framework that transforms video evaluation from passive benchmarking into a dynamic question-answering paradigm, yielding actionable feedback across diverse generative tasks.
    \item We formalize test-time scaling for video generation as a discrete, text-based optimization problem. By leveraging VLM-generated critiques as semantic gradients alongside a Global Selection and dynamic stopping mechanism, we iteratively correct visual flaws without requiring model weight access, effectively preventing semantic drift and ensuring efficiency.
    \item Extensive experiments demonstrate that \vqqa significantly outperforms state-of-the-art prompt optimization and sampling baselines across established benchmarks (T2V-CompBench~\citep{sun2025t2v}, VBench2~\citep{zheng2025vbench2}, VBench-I2V~\citep{huang2025vbench++}) using both open-weights and proprietary models~\citep{yang2024cogvideox, deepmind2025veo3}.
\end{itemize}

\begin{figure*}[t]
    \centering
    \includegraphics[width=0.95\linewidth]{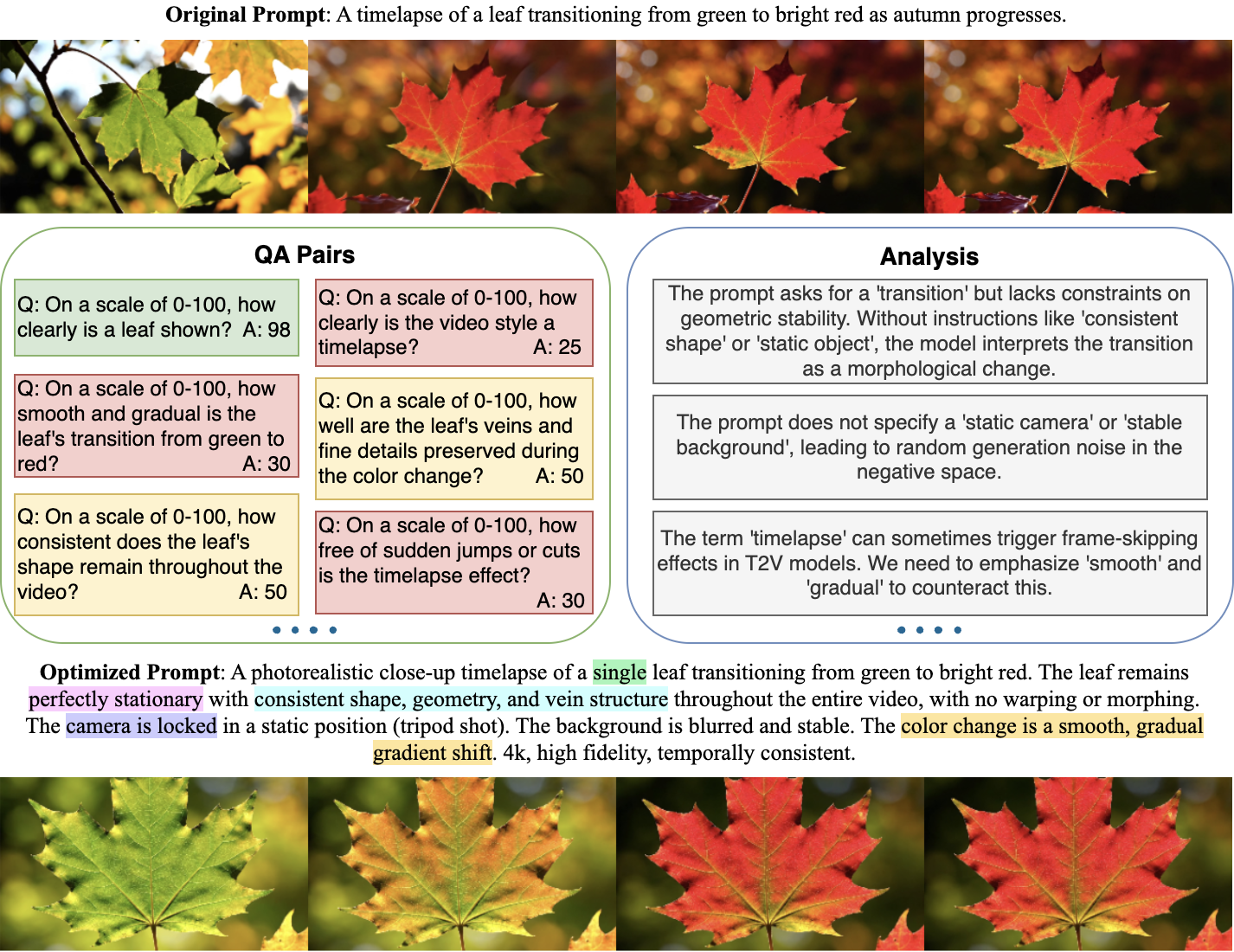}
    \caption{\textbf{Qualitative example of \vqqa iterative refinement process.} Utilizing low-scoring QA pairs to isolate visual flaws, \vqqa constructs an optimized prompt that successfully mitigates the localized artifacts in the next generation.}
\end{figure*}
\section{Related Work}
\label{related_work}

\subsection{Video Evaluation Frameworks}
The evaluation of generative video models has evolved from distribution-level metrics to semantic, agentic assessments. Early standard protocols relied on reference-based metrics like Fréchet Video Distance (FVD)~\citep{unterthiner2018towards} and Inception Score (IS)~\citep{salimans2016improved}. While extended by Fréchet Video Motion Distance (FVMD)~\citep{liu2024fr} to better capture temporal dynamics, these metrics correlate poorly with human perception at the instance level and fail to provide actionable feedback.

To bridge this gap, the field shifted toward VLM-based evaluation. Methods like CLIPScore~\citep{hessel2021clipscore} and BLIP-Score~\citep{li2022blip} measure frame-text consistency but lack temporal awareness. More recent approaches leverage multimodal large language models as judges: VQAScore~\citep{lin2024evaluating} computes probabilities of affirmative boolean answers, while T2VQA~\citep{kou2024subjective} and LMM-VQA~\citep{ge2025lmm} regress directly to Mean Opinion Scores (MOS). VideoScore2~\citep{he2025videoscore2} advances this by employing a ``think-before-you-score'' mechanism to generate Chain-of-Thought (CoT) rationales. However, its reasoning remains constrained to pre-defined axes, limiting its flexibility in diagnosing unique, instance-specific artifacts.

Most recently, comprehensive benchmark suites and modular pipelines~\citep{liu2023evalcrafter, huang2024vbench, huang2025vbench++, zheng2025vbench2, sun2025t2v} have decomposed quality into disentangled axes. While frameworks like Evaluation Agent~\citep{zhang-etal-2025-evaluation} employ human-like, multi-round evaluation, they remain primarily designed for passive performance benchmarking. In contrast, \vqqa dynamically generates context-dependent questions that serve as a direct language interface for downstream refinement.

\subsection{Prompt Optimization for Video Generation}
Prompt engineering has become essential for unlocking the capabilities of frozen generative models. Early text-based approaches like APE~\citep{zhou2022large} and Promptist~\citep{hao2023optimizing} utilized iterative search to align prompts with model preferences. In the visual domain, methods like Prompt-A-Video~\citep{ji2025prompt} and VPO~\citep{cheng2025vpo} adapt these techniques to text-to-video diffusion. VPO, for instance, optimizes prompts for harmlessness, accuracy, and helpfulness using Direct Preference Optimization (DPO)~\citep{rafailov2023direct}.

However, most of these methods operate in an ``open-loop'' fashion—optimizing prompts based on dataset-level priors rather than the specific visual artifacts of the \textit{current} generation. While self-correcting approaches are widely established in LLMs (e.g., Self-Refine~\citep{madaan2023self}, Reflexion~\citep{shinn2023reflexion}), systems for explicit self-critique and revision remain relatively underexplored in video generation. Recent exceptions include VideoAgent~\citep{soni2024videoagent} for robotic planning and VideoRepair~\citep{lee2024videorepair}, which employs a detect-and-patch'' strategy to fix localized misalignments. While effective for isolated errors, localized methods cannot address global inconsistencies like temporal flow. \vqqa overcomes this via a holistic ``closed-loop'' system, iteratively updating the entire prompt based on granular visual evidence to correct both local and global failures.

\subsection{Test-Time Scaling for Video Generation}
Scaling compute at inference time has proven effective for complex tasks. In LLMs, techniques like Tree-of-Thoughts (ToT)~\citep{yao2023tree} and Chain-of-Verification~\citep{dhuliawala2024chain} demonstrate that iterative reasoning boosts performance without retraining.

In video generation, inference-time scaling typically manifests as rejection sampling or trajectory search. Beyond the standard Best-of-N approach, Video-T1~\citep{liu2025videot1} formalizes test-time scaling as a trajectory search problem using verifiers, while VISTA~\citep{long2025vista} implements an agentic self-improving loop. Other approaches intervene directly in generative physics: Video-TTT~\citep{dalal2025one} applies gradient updates to RNN-based hidden states, while EvoSearch~\citep{he2025evosearch} mutates initial noise and intermediate latents to discover higher-quality trajectories.

However, existing approaches face distinct limitations. Gradient-based methods like Video-TTT and EvoSearch require \textit{white-box} access to model internals, rendering them incompatible with commercial APIs. Conversely, agentic frameworks like VISTA incur massive computational costs by requiring large candidate pools to identify improvements.

\vqqa proposes an alternative inspired by \textit{text-based optimization}. Building on TextGrad~\citep{yuksekgonul2024textgrad} and Feedback Descent~\citep{lee2025feedback}, which formalize backpropagation-like feedback via natural language, \vqqa treats the prompt as the optimization variable. By using VLM-guided feedback as \textit{semantic gradients}, \vqqa achieves precise adaptation and error correction strictly through a natural language interface, bypassing the need for weight access or exhaustive sampling.
\section{Methodology}
\subsection{Problem Formulation}

\subsubsection{Video Evaluation System}
Let $p$ denote a text prompt, $C = \{c_1, c_2, \dots, c_n\}$ a set of generation conditions (e.g., reference images), and $M$ a pre-trained video generation model. The sampled video $v$ is:
\begin{equation}
v = M(p, C)
\end{equation}

The objective of an evaluation system is to design a reward function $f$ that produces both a quantitative quality score $S_f$ and a qualitative linguistic rationale $R_f$ based on the provided inputs:
\begin{equation}
(S_f, R_f) = f(v, p, C)
\end{equation}

Given an external ground-truth evaluation system $g$ (e.g., human Mean Opinion Scores or established benchmark auto-raters), $f$ minimizes the discrepancy between its predicted score $S_f$ and the target score $S_g = g(v, p, C)$:
\begin{equation}
f^* = \argmin_f \mathbb{E} \left[ \Delta(S_g, S_f) \right]
\end{equation}

Crucially, $f$ requires \textbf{interpretability}; it must approximate the scalar score and generate human-interpretable reasoning $R_f$ to justify evaluations and facilitate downstream refinement~\citep{doshi2017towards}.

\subsubsection{Iterative Refinement via Test-Time Training (TTT)}
Function $f$ ultimately guides the optimization of prompt $p$. We formulate this as a Test-Time Training (TTT) task to find an optimal prompt $p^*$ from the prompt space $\mathcal{P}$:
\begin{equation}
p^* = \argmax_{p \in \mathcal{P}} g(M(p, C), p, C)
\end{equation}

Since $g$ is typically unknown and non-differentiable at inference time, we use $f \approx f^*$ as a proxy. This transforms video refinement into discrete prompt-space optimization process~\citep{yuksekgonul2024textgrad,lee2025feedback}. Rather than traditional gradient descent, which is ill-defined for discrete text, we use the reasoning component $R_{f,t}$ at step $t$ as a \textit{semantic gradient} to identify visual flaws. Letting $(S_{f,t}, R_{f,t}) = f(v_t, p_t, C)$, we define the iterative update rule as:
\begin{equation}
p_{t+1} = \text{VLM}(p_t, R_{f,t})
\end{equation}
where the VLM acts as a refinement operator, leveraging the critique $R_{f,t}$ to rectify flaws in the subsequent generation.

\subsection{The \vqqa Framework}
\begin{figure*}[t]
    \centering
    \includegraphics[width=\linewidth]{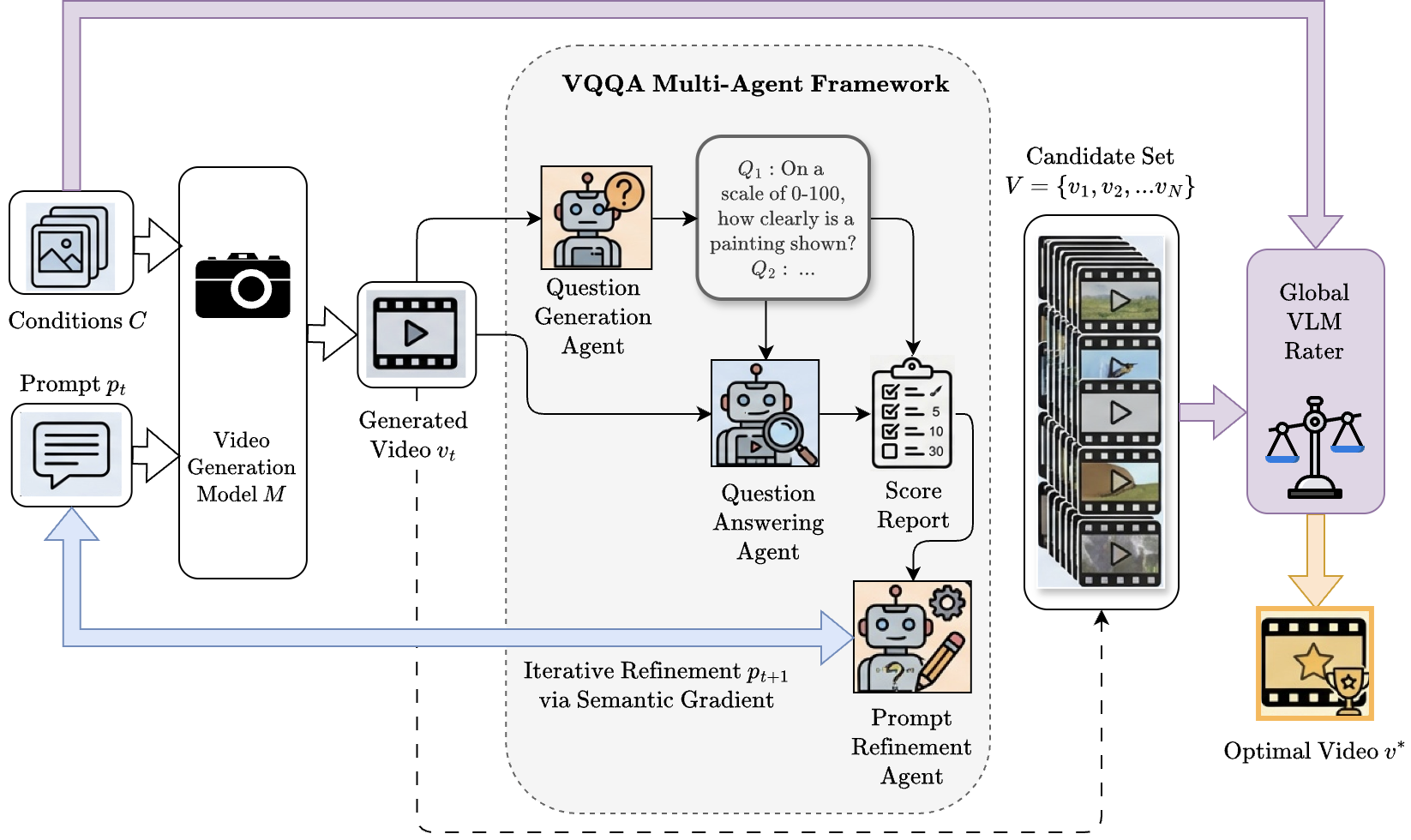}
    \caption{\textbf{The \vqqa framework:} Given generation conditions $C$ and a prompt $p_t$, the model $M$ produces a video $v_t$. The multi-agent framework uses a \textit{Question Generation (QG) agent} to formulate visual queries $Q$ and a \textit{Question Answering (QA) agent} to evaluate the video and produce a score report. These outputs inform the \textit{Prompt Refinement (PR) agent}, which uses semantic gradient to update the prompt for the next iteration. Finally, a \textit{Global VLM Rater} assesses the candidate set of generated videos against the original conditions to select the optimal video $v^*$.}
    \label{fig:vqqa_method}
\end{figure*}

\subsubsection{Multi-Agent Architecture}
As shown in \Cref{fig:vqqa_method}, \vqqa decomposes interpretable video evaluation and iterative refinement into three specialized agents:

\begin{itemize}
    \item \textbf{Question Generation (QG) Agent:} Analyzes the video $v$, prompt $p$, and conditions $C$ to dynamically generate a set of questions $Q$ across three 
    dimensions: \textit{Video-Prompt Alignment}, \textit{Visual Quality}, and \textit{Condition Fidelity} (when additional conditions $C$ are provided). This categorization directly mirrors the primary evaluation axes established by comprehensive benchmarks, ensuring robust and standardized coverage of critical failure modes.
    
    \item \textbf{Question Answering (QA) Agent:} Acts as the primary evaluator, inspecting video $v$ against questions $Q$ to assign normalized scores $s \in [0, 100]$ per question, constructing a detailed diagnostic map of critical visual flaws identified from the video.
    
    \item \textbf{Prompt Refinement (PR) Agent:} Synthesizes QA feedback into an optimized prompt $p_{t+1}$. By processing multiple low-scoring QA pairs (the semantic gradient), the agent formulates a revised prompt that concurrently mitigates these localized errors in the next iteration.
\end{itemize}

Complete implementation details, including the prompts, models and parameters, are provided in Appendix~\ref{sec:appendix_experiment_details}.

\subsubsection{Global Selection and Convergence}
The iterative refinement process is inherently stochastic, as it samples different trajectories within the latent space of the generator $M$. To ensure the framework identifies the optimal generation and terminates efficiently, we define both a global selection mechanism and a convergence criterion.

\paragraph{}\textbf{Global Selection} 
To prevent \textit{semantic drift}—where localized refinements cause deviation from overarching user intent—we employ a Global VLM Rater to perform a holistic, post-hoc evaluation of the candidate set $\mathcal{V} = \{v_1, v_2, \dots, v_N\}$. The rater assesses each candidate against the \textit{original} prompt $p_0$:
\begin{quote}\textit{On a scale of 0-100, how well does this video match `original prompt' (and `reference images')? \\ Respond with only a number between 0 and 100.}\end{quote}

The rater assigns a Global Score ($GS$) to each video, and the final output $v^*$ is selected as the candidate that maximizes alignment:
\begin{equation}
v^* = v_i, \quad \text{where } i = \argmax_{i \in \{1, \dots, N\}} GS(v_i, p_0, C)
\label{select_eq}
\end{equation}

This mechanism ensures that while the Prompt Refinement Agent explores variations to enhance visual quality, the final selection remains anchored to the user's primary goals. Unlike VQAScore~\citep{lin2024evaluating}, which relies on models fine-tuned specifically for text-visual alignment, our approach leverages the inherent visual reasoning capabilities of VLMs. This flexibility allows $GS$ to evaluate fidelity across multiple generation conditions without task-specific training.

\paragraph{}\textbf{Convergence Criterion}
To trade-off between inference cost and quality, we employ early stopping based on the running maximum Global Score $S^*_{t} = \max_{\tau \in \{1, \dots, t\}} GS(v_\tau)$. Refinement terminates at step $t$ if either condition is met:

\begin{enumerate}
    \item \textit{Target Satisfaction:} 
    The system generates a video that meets the ideal quality standard, denoted by a threshold $\gamma$ (e.g., $\gamma = 100$).
    \begin{equation}
        S^*_{t} \geq \gamma
    \end{equation}
    \item \textit{Performance Saturation:} The global maximum score stagnates over a ``patience'' window $k$, where improvement falls below a marginal $\epsilon$:
    \begin{equation}
        \Delta(S^*_t, S^*_{t-k})\leq \epsilon
    \label{saturation_eq}
    \end{equation}
\end{enumerate}

This ensures that the framework halts immediately upon achieving an optimal result or when additional compute yields no additional improvement.
\section{Experiments}
\label{sec:exp}

\subsection{Experimental Setup}
\subsubsection{Tasks}

We evaluate the \vqqa framework on two video generation tasks:

\paragraph{Text-to-Video (T2V):}Given a text prompt $p$, the objective is to generate a video $v = M(p)$ that accurately reflects the semantic and temporal content of $p$.

\paragraph{Image-to-Video (I2V):}Given a prompt $p$ and reference images $C = \{i_1, \dots, i_m\}$, the goal is to synthesize a video $v = M(p, C)$ adhering to the textual prompt while maintaining high visual fidelity to the reference image(s).

\paragraph{} Unlike traditional pipelines hard-coded for specific modalities, \vqqa acts as a task-agnostic optimizer. Its agents dynamically adapt visual queries to the provided condition set (text-only or text+images). This versatility allows us to seamlessly apply the exact same iterative refinement process to both paradigms without architectural modifications.

\begin{table*}[tbp]
\caption{\textbf{T2V-CompBench evaluation results on CogVideoX-5B}. Performance is reported using $N=5$ for Best-of-$N$ strategy, and 4 optimization rounds for \vqqa. Vanilla generation results are obtained from the official leaderboard. Best and second-best scores are highlighted in green and yellow. All numbers are percentages.}
\label{tab:compbench_cogvideox}
\centering
\resizebox{\linewidth}{!}{
\begin{tabular}{llcccccccc} 
\toprule


\multicolumn{2}{c}{\textbf{Method}} & \textbf{Consist-attr} & \textbf{Dynamic-attr} & \textbf{Spatial} & \textbf{Motion} & \textbf{Action} & \textbf{Interaction} & \textbf{Numeracy} & \textbf{AVG} \\
\midrule
\multicolumn{2}{l}{\textit{Vanilla Generation}} \\
& CogVideoX-5B~\citep{yang2024cogvideox} & 61.64 & 2.19 & 51.72 & 26.58 & 53.33 & 60.69 & 37.06 & 41.89 \\
\midrule
\multicolumn{2}{l}{\textit{Prompt Optimizer}} \\
& VPO~\citep{cheng2025vpo} & 76.43 & 4.52 & 54.18 & 26.07 & 64.09 & \cellcolor{second_color}68.83 & 45.72 & 48.55 \\
\midrule
\multicolumn{2}{l}{\textit{Best-of-N $\And$}} \\
& VQAScore~\citep{lin2024evaluating} & 80.83 & 2.48 & 56.62 & 26.24 & 64.31 & 65.96 & 44.48 & 48.70 \\
& VideoScore2~\citep{he2025videoscore2} & 80.86 & 1.99 & 56.90 & 26.82 & 58.24 & 61.54 & 39.53 & 46.55 \\
\cmidrule{2-10}
\multicolumn{2}{l}{\textit{Best-of-N $\And$ VLM-Rating}} \\
& GPT-4o~\citep{hurst2024gpt} & 80.17 & 2.00 & 56.36 & 27.29 & 60.59 & 66.06 & 42.19 & 47.81 \\
& Gemini-3-Pro~\citep{google_gemini3_blog_2025} & 79.00 & 2.34 & 58.36 & \cellcolor{best_color}28.63 & 60.31 & 58.97 & 43.42 & 47.29 \\
\midrule
\multicolumn{2}{l}{\textit{\vqqa}} \\
& GPT-4o~\citep{hurst2024gpt} & \cellcolor{second_color}82.96 & \cellcolor{second_color}4.88 & \cellcolor{second_color}58.73 & \cellcolor{second_color}28.51 & \cellcolor{second_color}66.72 & \cellcolor{best_color}69.57 & \cellcolor{second_color}47.75 & \cellcolor{second_color}51.30 \\
& Gemini-3-Pro~\citep{google_gemini3_blog_2025} & \cellcolor{best_color}84.58 & \cellcolor{best_color}7.92 & \cellcolor{best_color}66.03 & 27.73 & \cellcolor{best_color}68.50 & 68.54 & \cellcolor{best_color}50.91 & \cellcolor{best_color}\textbf{53.46} \\
\bottomrule
\end{tabular}%
}
\end{table*}
\begin{table*}[tbp]
\caption{\textbf{VBench2 evaluation results on CogVideoX-5B}. Performance is reported with $N=5$ (Best-of-$N$) and 4 optimization rounds (\vqqa). All numbers are percentages.}
\label{tab:vbench2_cogvideox}
\centering
\resizebox{\linewidth}{!}{
\begin{tabular}{llcccccc} 
\toprule

\multicolumn{2}{c}{\textbf{Method}} & \textbf{Creativity} & \textbf{Commonsense} & \textbf{Controllability} & \textbf{Human Fidelity} & \textbf{Physics} & \textbf{Total Score} \\

\midrule
\multicolumn{2}{l}{\textit{Vanilla Generation}} \\
& CogVideoX-5B~\citep{yang2024cogvideox} & 42.99 & 54.19 & 23.81 & 80.13 & 38.57 & 41.98\\
\midrule
\multicolumn{2}{l}{\textit{Prompt Optimizer}} \\
& VPO~\citep{cheng2025vpo} &  44.44 & 54.19 & 24.79 & \cellcolor{best_color}84.45 & 39.01 & 43.34 \\
\midrule
\multicolumn{2}{l}{\textit{Best-of-N $\And$}} \\
& VQAScore~\citep{lin2024evaluating} & 51.51 & 57.92 & \cellcolor{second_color}30.96 & 79.85 & 42.49 & 46.95 \\
& VideoScore2~\citep{he2025videoscore2} & 45.92 & 55.05 & 26.92 & 78.09 & 35.46 & 43.97  \\
\cmidrule{2-8}
\multicolumn{2}{l}{\textit{Best-of-N $\And$ VLM-Rating}} \\
& GPT-5.1~\citep{singh2025openai} & 48.34 & 56.49 & 29.81 & 78.83 & 33.88 & 45.28  \\
& GPT-4o~\citep{hurst2024gpt} & 48.19 & 57.06 & 28.25 & 80.05 & 41.03 & 46.15 \\
& Gemini-3-Pro~\citep{google_gemini3_blog_2025} & 47.96 & 57.64 & 28.79 & 79.69 & 31.05 & 44.76  \\
\midrule
\multicolumn{2}{l}{\textit{\vqqa}} \\
& GPT-4o~\citep{hurst2024gpt} & \cellcolor{second_color}55.47 & \cellcolor{second_color}57.92 & 30.41 & 79.95 & \cellcolor{second_color}46.92 & \cellcolor{second_color}48.18  \\
& Gemini-3-Pro~\citep{google_gemini3_blog_2025} & \cellcolor{best_color}54.85 & \cellcolor{best_color}58.21 & \cellcolor{best_color}31.50 & \cellcolor{second_color}81.23 & \cellcolor{best_color}54.26 & \cellcolor{best_color}\textbf{50.41} \\
\bottomrule
\end{tabular}%
}
\end{table*}

\subsubsection{Baselines}
We compare our method against established optimization frameworks and stochastic search strategies using various scoring functions:

\begin{itemize}
\item \textbf{Video Prompt Optimization (VPO)}~\citep{cheng2025vpo}: A primary baseline for prompt refinement. VPO is a two-stage framework optimizing for harmlessness, accuracy, and helpfulness. It refines prompts to enhance both generation quality and safety.

\item \textbf{Best-of-N (BoN)}~\citep{stiennon2020learning}: A stochastic search baseline sampling $N$ candidates $\{v_1, \dots, v_N\}$ from the initial prompt $p_0$, selecting the optimal video via a reward function. We implement BoN with three distinct scoring mechanisms:

\begin{itemize}
    \item \textit{BoN with VQAScore}~\citep{lin2024evaluating}: VQAScore is an automated metric based on fine-tuned Visual Question Answering (VQA) models. It measures alignment by averaging confidence scores of boolean questions (e.g., ``Does this video show [text]?'') across video frames.
    \item \textit{BoN with VideoScore2}~\citep{he2025videoscore2}: VideoScore2 is a regression-based reward model trained on large-scale human preference data. Unlike frame-averaging metrics, it uses a spatio-temporal representation to directly predict human-like quality and alignment scores for an entire video sequence.
    \item \textit{BoN with VLM-Rating}: Prompts multimodal VLMs to act as judges, rating video-condition alignment on a 0--100 scale.
\end{itemize}
\end{itemize}

\subsubsection{Benchmarks}
We evaluate our method on the following benchmarks:

\begin{itemize}
    \item \textbf{T2V-CompBench}~\citep{sun2025t2v}: A text-to-video benchmark comprising 1,400 prompts across seven categories. It is designed to evaluate compositional generation, measuring how effectively models integrate multiple objects, actions, and attributes into coherent, temporally consistent videos.
    
    \item \textbf{VBench2}~\citep{huang2024vbench}: A text-to-video evaluation suite focused on measuring intrinsic faithfulness. VBench2 uses a multi-agent VLM system to evaluate five high-level dimensions: Human Fidelity, Controllability, Creativity, Physics, and Commonsense. This framework enables systematic assessment of a model’s ability to reduce complex physical and anatomical hallucinations.    
    \item \textbf{VBench-I2V}~\citep{huang2025vbench++}: An extension of VBench for image-to-video generation. It evaluates how faithfully a generated video aligns with a reference image, assessing temporal consistency, motion magnitude, and the preservation of the image’s identity and semantic details.
\end{itemize}

\subsection{Results}
\label{sec:results}

\subsubsection{T2V-CompBench} 
\Cref{tab:compbench_cogvideox} shows our \vqqa framework consistently outperforming both the vanilla generation and all baselines on T2V-CompBench. Using Gemini-3-Pro, \vqqa achieves the highest average score (53.46\%), delivering an absolute improvement of \textbf{+11.57\%} over vanilla generation and \textbf{+4.76\%} over the strongest baseline (VQAScore). \vqqa with GPT-4o secures the second-best average (51.30\%), showing the superiority of our approach regardless of the underlying VLM. The iterative refinement process effectively resolves compositional errors, yielding substantial absolute gains across all categories, especially in consistent-attribute (\textbf{+22.94\%}), spatial understanding (\textbf{+14.31\%}), and numeracy (\textbf{+13.85\%}).

\subsubsection{VBench2} 
Results on the VBench2 benchmark (\Cref{tab:vbench2_cogvideox}) further validate our approach. The Gemini-3-Pro and GPT-4o variants of \vqqa achieve the highest (50.41\%) and second-highest (48.18\%) total scores, respectively. \vqqa with Gemini-3-Pro provides a \textbf{+8.43\%} absolute increase over the vanilla baseline and surpasses the best competing method (VQAScore) by \textbf{+3.46\%}. 

\subsubsection{VBench-I2V} 
Evaluation results on the VBench-I2V benchmark (\Cref{tab:vbenchi2v_cogvideox}) further demonstrate the versatility of the \vqqa framework on the I2V task. Despite the high saturation of this benchmark, \vqqa with Gemini-3-Pro achieves the highest performance across all evaluated axes, improving upon vanilla generation by \textbf{+1.24\%} and the strongest Best-of-N baseline by \textbf{+0.23\%}. Furthermore, \vqqa exhibits remarkable efficiency, requiring an average of only \textbf{1.6} iterations to satisfy the task's stopping criterion.

\subsection{Analysis}
\label{sec:analysis}

\begin{table*}[tbp]
\caption{\textbf{VBench-I2V evaluation results on CogVideoX-5B-I2V}. Performance is reported with $N=5$ (Best-of-$N$) and 4 optimization rounds (\vqqa). Vanilla Generation results are obtained from the  leaderboard. All numbers are percentages.}
\label{tab:vbenchi2v_cogvideox}
\centering
\resizebox{\linewidth}{!}{
\begin{tabular}{llcccccc} 
\toprule

\multicolumn{2}{c}{\textbf{Method}} & \makecell{\textbf{I2V} \\ \textbf{Background}} & \makecell{\textbf{Background} \\ \textbf{Consistency}} & \makecell{\textbf{I2V} \\ \textbf{Subject} } & \makecell{\textbf{Subject} \\ \textbf{Consistency}} & \makecell{\textbf{Motion} \\ \textbf{Smoothness}} & \makecell{\textbf{Avg} \\ \textbf{Score}} \\
\midrule
\multicolumn{2}{l}{\textit{Vanilla Generation}} \\
& CogVideoX-5B-I2V~\citep{yang2024cogvideox} & 96.74 & 96.42 & 97.19 & 94.34 & \cellcolor{second_color}98.40 & 96.62 \\
\midrule
\multicolumn{2}{l}{\textit{Best-of-N $\And$}} \\
& VQAScore~\citep{lin2024evaluating} & 98.96 & 97.19 & 97.61 & 95.33 & 98.29 & 97.48 \\
\cmidrule{2-8}
\multicolumn{2}{l}{\textit{Best-of-N $\And$ VLM-Rating}} \\
& Gemini-3-Pro~\citep{google_gemini3_blog_2025} & \cellcolor{second_color}99.03 & \cellcolor{second_color}97.37 & \cellcolor{second_color}97.77 & \cellcolor{second_color}95.68 & 98.32 & \cellcolor{second_color}97.63 \\
\midrule
\multicolumn{2}{l}{\textit{\vqqa}} \\
& Gemini-3-Pro~\citep{google_gemini3_blog_2025} & \cellcolor{best_color}99.08 & \cellcolor{best_color}97.40 & \cellcolor{best_color}98.09 & \cellcolor{best_color}96.29 & \cellcolor{best_color}98.44 & \cellcolor{best_color}\textbf{97.86} \\
\bottomrule
\end{tabular}%
}
\end{table*}
\begin{table}[tbp]
    \caption{\textbf{Quality of the visual flaw identification process (Judged by Gemini-3-Flash)}. Both methods use Gemini-3-Pro as the backbone. \vqqa Score threshold for E2E-Recall measurement is set to 60. All numbers are percentages.}
    \label{tab:ablation_question_quality}
    \centering
    \begin{tabular}{l ccc}
        \toprule
        \textbf{Method} & \textbf{Precision} & \textbf{Q-Recall} & \textbf{E2E-Recall} \\
        \midrule
        VLM Direct Analysis & 99.94 & -- & 70.18 \\
        \vqqa                & 99.28 & 96.55 & 82.08 \\
        \bottomrule
    \end{tabular}
\end{table}

\subsubsection{Quality of Generated Questions}
To evaluate the coverage and effectiveness of \vqqa's generated questions, we use the VideoFeedback2~\citep{he2025videoscore2} test split. We construct a ground-truth (GT) set of visual flaws by prompting GPT-5.2~\citep{singh2025openai} to extract discrete problems from the dataset's reasoning trajectories. A judge model maps each GT problem to the specific \vqqa questions designed to detect it. We evaluate recall at two levels: (1) \textit{Question Recall (Q-Recall)}, the percentage of GT problems covered by at least one generated question; and (2) \textit{End-to-End (E2E) Recall}, the percentage of GT problems where the corresponding \vqqa question correctly receives a sub-threshold score from the QA agent.

We measure precision based on the generated questions' relevance to the input video and prompt, assessed via binary classification by a judge model. We prioritize relevance over strict GT mapping for two reasons: (i) the GT set, derived from Claude-4-Sonnet~\citep{anthropic2025claude4} trajectories, may be incomplete; and (ii) \vqqa proactively probes for \textit{potential} visual artifacts. Consequently, a contextually relevant question remains valuable even if no flaw is ultimately confirmed.

Finally, we compare \vqqa against a zero-shot VLM baseline that directly identifies visual flaws. As \Cref{tab:ablation_question_quality} demonstrates, while both methods maintain near-perfect precision (>99\%), \vqqa yields a significant \textbf{11.9\%} improvement in \textbf{E2E-Recall} over the baseline. This substantial gain in recall means our system captures a considerably wider array of visual artifacts, delivering the robust feedback necessary to effectively correct errors and improve subsequent video generations. Qualitative analysis indicates that the marginal precision deficits in both methods primarily stem from infrequent VLM hallucinations during evaluation.

\subsubsection{Convergence and Efficiency}
\begin{figure*}[t]
    \centering
    \begin{subfigure}[b]{0.48\linewidth}
        \centering
        \includegraphics[width=0.8\linewidth]{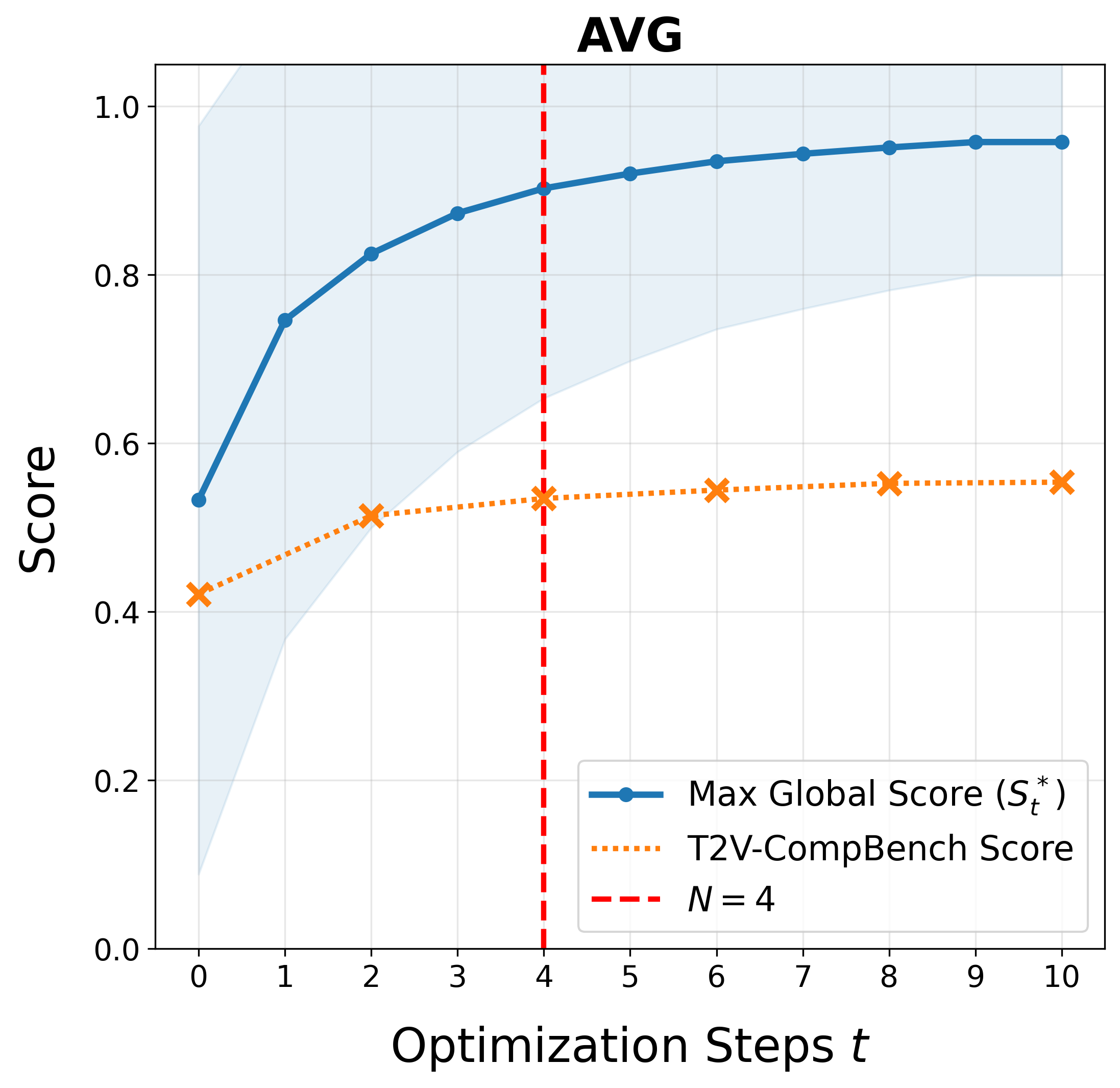}
        \caption{\vqqa performance across iterations}
        \label{fig:convergence_avg}
    \end{subfigure}
    \hfill
    \begin{subfigure}[b]{0.48\linewidth}
        \centering
        \includegraphics[width=0.8\linewidth]{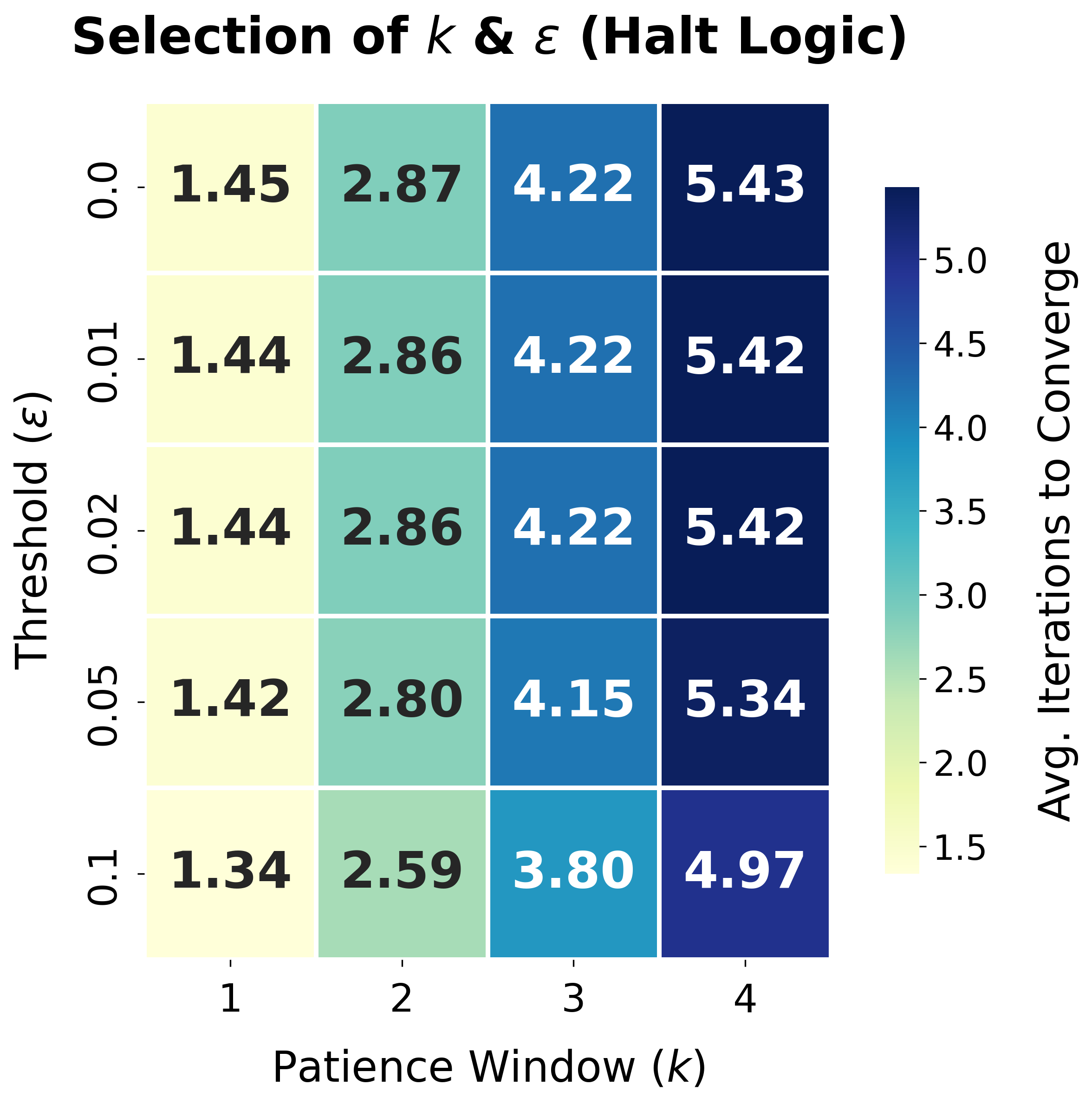}
        \caption{Convergence parameter sensitivity}
        \label{fig:convergence_heatmap}
    \end{subfigure}
    
    \caption{\textbf{Convergence analysis of \vqqa on T2V-CompBench.} Evaluations are performed for CogVideoX-5B generations using Gemini-3-Pro. 
    \textbf{(a)} Correlations between the maximum global score $S^*_t$ and the T2V-CompBench metric across optimization steps. The blue shaded region indicates $\pm 1$ standard deviation of $S^*_t$ across the 1400 evaluated samples. The detailed performance breakdown for each individual category is provided in Appendix~\ref{subsec:long_horizon_runs}.  
    \textbf{(b)} Sensitivity analysis of average iterations to converge given the patience window $k$ and saturation threshold $\epsilon$ (\Cref{saturation_eq}).}
    \label{fig:convergence_analysis}
\end{figure*}

To evaluate the efficiency of the proposed stopping criterion (\Cref{saturation_eq}), we analyze optimization trajectories over a horizon of $N=10$ steps. As illustrated in \Cref{fig:convergence_analysis}, both the agent’s internal score $S^*_t$ and the external T2V-CompBench metrics exhibit rapid gains in the initial few iterations, reaching a saturation plateau shortly thereafter. 

The sensitivity analysis in \Cref{fig:convergence_heatmap} demonstrates that for a standard patience window $k=3$, the algorithm converges within $3.80$ to $4.22$ iterations across various thresholds. This aligns with the visual evidence in \Cref{fig:convergence_avg}, where the red dashed line (Iter=4) indicates the majority of performance gains are captured within the first four rounds. These results confirm that our dynamic termination mechanism effectively prevents redundant inference steps by halting once the semantic gradient vanishes, thereby maximizing computational efficiency without sacrificing the video quality gain accumulated through iterations.

\subsection{Ablation Studies}
\label{sec:ablation}

\begin{table*}[tbp]
\caption{\textbf{T2V-CompBench evaluation results on Veo3.1}. Performance is reported with $N=5$ (Best-of-$N$) and 4 optimization rounds (VQQA). All numbers are percentages.}
\label{tab:compbench_veo}
\centering
\resizebox{\linewidth}{!}{
\begin{tabular}{llcccccccc} 
\toprule


\multicolumn{2}{c}{\textbf{Method}} & \textbf{Consist-attr} & \textbf{Dynamic-attr} & \textbf{Spatial} & \textbf{Motion} & \textbf{Action} & \textbf{Interaction} & \textbf{Numeracy} & \textbf{AVG} \\
\midrule
\multicolumn{2}{l}{\textit{Vanilla Generation}} \\
& Veo3.1~\citep{deepmind2025veo3} & 76.45 & 25.70 & 59.76 & 51.18 & 71.81 & 62.87 & 43.77 & 55.93 \\
\midrule
\multicolumn{2}{l}{\textit{Best-of-N $\And$}} \\
& VQAScore~\citep{lin2024evaluating} & 78.21 & 26.13 & \cellcolor{second_color}63.76 & 52.03 & \cellcolor{best_color}74.83 & \cellcolor{second_color}66.99 & \cellcolor{second_color}54.59 & 59.51 \\
& VideoScore2~\citep{he2025videoscore2} & 78.19 & 24.76 & 60.99 & 54.49 & 73.52 & 65.98 & 46.7 & 57.80 \\
\cmidrule{2-10}
\multicolumn{2}{l}{\textit{Best-of-N $\And$ VLM-Rating}} \\
& Gemini-3-Pro~\citep{google_gemini3_blog_2025} & \cellcolor{second_color}78.70 & \cellcolor{best_color}27.56 & 62.98 & \cellcolor{second_color}55.61 & 73.63 & 65.33 & 49.25 & 59.01 \\
\midrule
\multicolumn{2}{l}{\textit{\vqqa}} \\
& Gemini-3-Pro~\citep{google_gemini3_blog_2025} & \cellcolor{best_color}82.00 & \cellcolor{second_color}26.44 & \cellcolor{best_color}65.73 & \cellcolor{best_color}59.70 & \cellcolor{second_color}74.31 & \cellcolor{best_color}68.02 & \cellcolor{best_color}56.47 & \cellcolor{best_color}\textbf{61.81}\\
\bottomrule
\end{tabular}%
}
\end{table*}
\subsubsection{Generalizability to Proprietary Models}
To assess the model-agnostic nature of \vqqa, we applied our prompt optimization pipeline to the proprietary Google \textbf{Veo~3.1} model~\citep{deepmind2025veo3}. As shown in \Cref{tab:compbench_veo}, \vqqa remains highly effective even with Veo~3.1's mandatory internal prompt optimization enabled. After four rounds, \vqqa improved the overall T2V-CompBench average score by an absolute margin of \textbf{+5.88\%} (from 55.93\% to 61.81\%), outperforming the Best-of-N approaches using VQAScore (\textbf{+3.58\%}) and Gemini-3-Pro (\textbf{+3.08\%}). Furthermore, \vqqa achieves the highest performance across the majority of compositional dimensions.

\begin{figure*}[t]
    \centering
    \includegraphics[width=\linewidth]{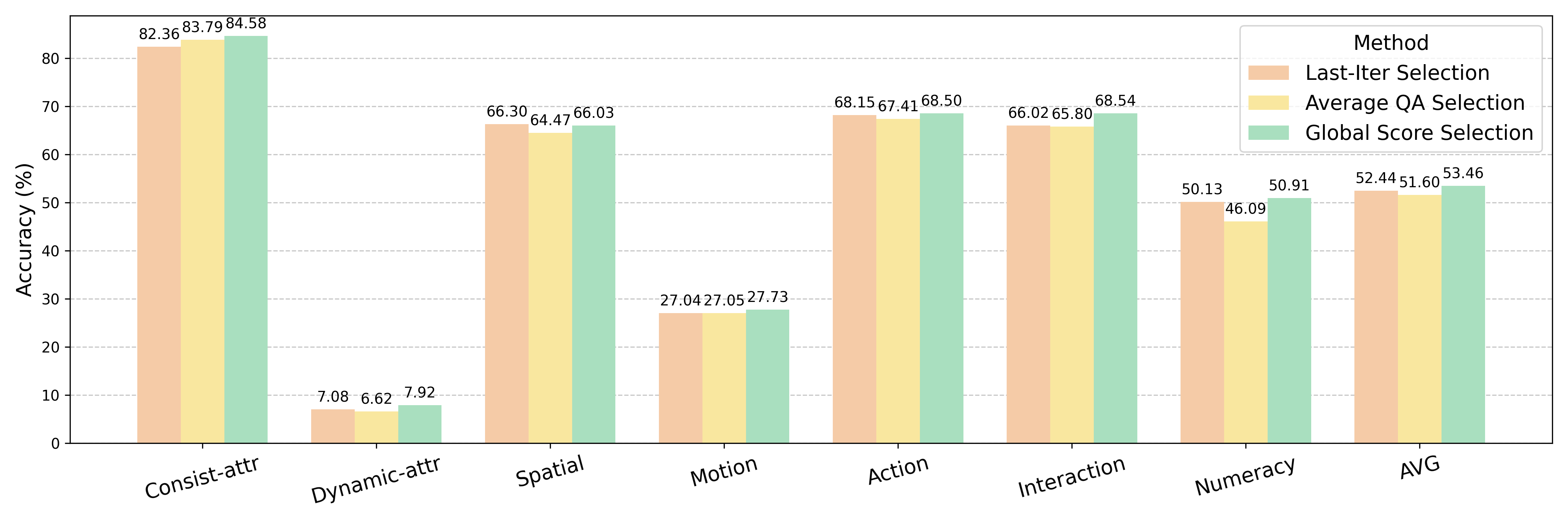}
    \caption{\textbf{Ablation Study on Global Selection Mechanism}. The number of prompt optimization rounds for \vqqa is fixed at N = 4.}
    \label{fig:ablation_study_selection_methods}
\end{figure*}

\subsubsection{Global Selection Mechanism}
We investigate the contribution of the Global Selection (\Cref{select_eq}) by comparing the selection function against two variants:

\begin{enumerate}
    \item \textbf{Last-Iter Selection ($v_N$):} We disable the Global Rater and simply accept the output of the final iteration. A performance drop here indicates ``semantic drift,'' where the Prompt Refinement Agent optimizes for localized visual details at the expense of original user intent. As shown in \Cref{fig:ablation_study_selection_methods}, omitting global selection decreases the overall average score by 1.02\%, confirming that unanchored refinement drifts from primary constraints.
    
    \item \textbf{Average-QA Selection:} We replace the Global Score $GS$ with the average score derived from the QA Agent's internal questions. This tests whether a holistic, post-hoc evaluation supersedes the aggregated granular scores used during optimization. This variant underperforms the proposed global selection by an average margin of 1.86\%. This confirms that while granular QA scores provide effective semantic gradients for localized refinement, their unweighted aggregation fails to capture the ``gestalt alignment''~\citep{wagemans2012century1,wagemans2012century2} required for optimal video selection.
\end{enumerate}

Overall, \Cref{fig:ablation_study_selection_methods} demonstrates the necessity of decoupling the iterative feedback loop from the final selection criterion. Because prompt refinement targets specific visual bottlenecks, it does not guarantee monotonic improvement in holistic text-to-video alignment at every step. By employing the Global Score to evaluate the candidate pool against the original prompt, the proposed mechanism successfully mitigates semantic drift. This macro-level perspective ensures the final output achieves the optimal balance across critical dimensions, outperforming alternative selection strategies.

\begin{table*}[tbp]
\caption{\textbf{Ablation Study on GlobalScore in the loop}. Number of \vqqa optimization rounds is fixed at N = 4. Best and second-best average scores across models for each strategy are highlighted in green and yellow. All numbers are percentages.}
\label{tab:ablation_gs_in_the_loop}
\centering
\resizebox{\linewidth}{!}{
\begin{tabular}{llcccccccc} 
\toprule


\multicolumn{2}{c}{\textbf{Method}} & \textbf{Consist-attr} & \textbf{Dynamic-attr} & \textbf{Spatial} & \textbf{Motion} & \textbf{Action} & \textbf{Interaction} & \textbf{Numeracy} & \textbf{AVG} \\

\midrule
\multicolumn{2}{l}{\textit{\vqqa \textbf{GS-in-the-loop}}} \\
& Gemini-3-Pro~\citep{google_gemini3_blog_2025} & 81.77 & 8.40 & 67.05 & 27.94 & 68.15 & 70.83 & 50.69 & 53.55 \\
& Gemini-2.5-Pro~\citep{comanici2025gemini} & 83.92 & 6.32 & 64.47 & 28.19 & 66.33 & 72.69 & 46.67 & 52.66 \\
& AVG & \cellcolor{second_color}82.85 & \cellcolor{second_color}7.36 & \cellcolor{second_color}65.76 & \cellcolor{best_color}28.07 & \cellcolor{second_color}67.24 & \cellcolor{best_color}71.76 & \cellcolor{second_color}48.68 & \cellcolor{second_color}53.11 \\
\midrule
\multicolumn{2}{l}{\textit{\vqqa 
\textbf{Standard}}} \\
& Gemini-3-Pro~\citep{google_gemini3_blog_2025} & 84.58 & 7.92 & 66.03 & 27.73 & 68.50 & 68.54 & 50.91 & 53.46 \\
& Gemini-2.5-Pro~\citep{comanici2025gemini} & 83.43 & 7.08 & 65.99 & 27.76 & 68.17 & 72.37 & 53.66 & 54.07 \\
& AVG & \cellcolor{best_color}84.01 &\cellcolor{best_color}7.50 & \cellcolor{best_color}66.01 & \cellcolor{second_color}27.75 & \cellcolor{best_color}68.34 & \cellcolor{second_color}70.46 & \cellcolor{best_color}52.29 & \cellcolor{best_color}\textbf{53.77} \\
\bottomrule
\end{tabular}%
}
\end{table*}

\subsubsection{Prompt Refinement: GS-in-the-loop (or not?)} 
To evaluate if the global score $GS$ can further impact the optimization process, we tried one variation of our method, which is injecting the global score into the Prompt Refinement (PR) agent's context history, analyzing whether exposing the PR agent to this high-level metric aids convergence or introduces noise that hampers specific visual corrections. 

As \Cref{tab:ablation_gs_in_the_loop} shows, incorporating $GS$ during iterative refinement degrades average generation quality (53.11\% vs. the standard 53.77\%). Specifically, excluding $GS$ from the PR context better resolves fine-grained compositional errors, outperforming the GS-in-the-loop variant in attribute consistency (84.01\% vs. 82.85\%), spatial understanding (66.01\% vs. 65.76\%), and numeracy (52.29\% vs. 48.68\%).

This indicates that providing a holistic metric during refinement introduces optimization noise. When exposed to $GS$, the agent tends to become distracted from the localized, specific bottlenecks identified by the granular QA pairs. It attempts to prematurely optimize the global score rather than sequentially resolving isolated visual errors. Therefore, strictly decoupling the granular feedback loop (used for targeted refinement) from the global score (used strictly for final candidate selection) yields the most robust compositional improvements.

\section{Conclusion}
\label{sec:conclusion}
In this paper, we introduced \vqqa, a novel multi-agent framework that transforms the passive evaluation of video generation models into an active, closed-loop refinement process. By dynamically generating visual questions and leveraging Vision-Language Model critiques as semantic gradients, \vqqa enables precise, iterative prompt optimization through a black-box natural language interface. Extensive evaluations across both T2V and I2V tasks demonstrate that \vqqa yields significant improvements over strong baselines requiring very few iterations, generalizing seamlessly across open-weights and proprietary models. Ultimately, \vqqa provides a scalable, task-agnostic solution for aligning visual generative models with complex human intent, paving the way for more controllable and interpretable AI-driven content creation.

\bibliography{main}
\clearpage

\appendix
\section{Limitations}

While \vqqa demonstrates strong performance, its efficacy is inherently bounded by its underlying foundation models. As a black-box optimizer, it relies on Vision-Language Models (VLMs) for reasoning and base video models for generation. Consequently, it cannot rectify fundamental architectural flaws, synthesize out-of-distribution concepts, or entirely eliminate noisy semantic gradients from VLM hallucinations. However, its model-agnostic design ensures these constraints will naturally diminish as foundation models advance. Furthermore, the iterative multi-agent loop requires sequential generation and querying, resulting in higher inference latency than parallelizable methods like Best-of-$N$. Nevertheless, \vqqa justifies this sequential compute cost by achieving rapid convergence and superior alignment with complex user intent.
\section{Inference Cost}

We analyze the number of VLM calls required by our pipeline compared to the baseline methods. Let $N$ denote the number of sampled candidates in the Best-of-$N$ strategy, and $T$ denote the number of optimization iterations in \vqqa.

As detailed in Table~\ref{tab:inference_cost}, standard Best-of-$N$ approaches require $N$ parallel VLM calls to evaluate the pre-generated candidate pool. In contrast, \vqqa actively forms a semantic gradient at each step, requiring multiple VLM calls per iteration: Question Generation (2 calls for T2V, plus $k$ calls for $k$ reference images in I2V settings), Question Answering (1 call), and Prompt Refinement (1 call). This results in $(4+k)T$ generation-phase calls. The Global Rater then evaluates the $T+1$ generated candidates (the initial generation plus the $T$ optimized outputs), adding $T+1$ evaluation calls. 

While the theoretical maximum number of VLM calls for \vqqa is $(5+k)T+1$, our method converges rapidly in practice due to the dynamic stopping criterion. On the T2V-CompBench dataset ($k=0$), \vqqa converges after an average of $T = 1.245$ optimization rounds using the Gemini-3-Pro~\citep{google_gemini3_blog_2025} model. By substituting this average into our cost formulation, the expected total number of VLM calls for \vqqa is approximately $7.23$. This demonstrates that \vqqa achieves its significant improvements in visual quality and prompt alignment with an inference footprint comparable to a standard Best-of-5 ($N=5$) baseline.

\begin{table}[h!]
\centering
\caption{Comparison of VLM inference costs per prompt. For \vqqa, $k$ denotes the number of reference images in I2V settings ($k=0$ for T2V). When compared against a Best-of-5 ($N=5$) baseline, \vqqa requires an average of $T = 1.245$ optimization rounds to converge on T2V-CompBench (capped at a maximum of 4 optimization rounds).}
\label{tab:inference_cost}
\begin{tabular}{ccccc}
\toprule
\multirow{2}{*}{Method} & \multicolumn{3}{c}{\#VLM Calls} & \multirow{2}{*}{\shortstack{T2V-CompBench\\Total\\($N=5, k=0$)}} \\
\cmidrule(lr){2-4}
 & Generation & Evaluation & Total & \\
\midrule
Vanilla Generation       & 0        & 0     & 0          & 0 \\
Best-of-$N$ (VLM-Rating) & 0        & $N$   & $N$        & 5 \\
\vqqa (Ours)             & $(4+k)T$ & $T+1$ & $(5+k)T+1$ & $\sim 7.23$ \\
\bottomrule
\end{tabular}
\end{table}

\section{Experiment Details}
\label{sec:appendix_experiment_details}
\subsection{Models}
\subsubsection{Vision-Language Models (VLMs)}
\begin{itemize}
    \item \textbf{Gemini API:} We accessed the Gemini model (\texttt{gemini-3-pro-preview} and \texttt{gemini-3-flash\\-preview})~\citep{google_gemini3_blog_2025} via the Google Cloud Vertex AI platform.
    \item \textbf{OpenAI API:} We utilized OpenAI models (\texttt{gpt-4o-2024-08-06}~\citep{hurst2024gpt}, \texttt{gpt-\\5.1-2025-11-13} and \texttt{gpt-5.2-2025-12-11}~\citep{singh2025openai}) accessed via the official OpenAI API.
\end{itemize}

\subsubsection{Video Generation Models}
\begin{itemize}
    \item \textbf{CogVideoX-5B:}~\citep{yang2024cogvideox} Accessed via the Hugging Face \texttt{diffusers} library (\texttt{THUDM/\\CogVideoX-5b}). Generation parameters were set to 50 inference steps, 41 frames, and a guidance scale of 6.0.
    \item \textbf{Veo 3.1:}~\citep{deepmind2025veo3} Accessed via the Google Cloud Vertex AI API (\texttt{veo-3.1-gene\\rate-preview}). Videos were generated with a 16:9 aspect ratio using default parameters. Generations that triggered safety filters were replaced with blank videos to maintain batch consistency.
    \item \textbf{CogVideoX-5B-I2V:}~\citep{yang2024cogvideox} Accessed via the \texttt{diffusers} library (\texttt{THUDM/CogVi\\deoX-5b-I2V}). Generation parameters were set to 50 inference steps, 49 frames, and a guidance scale of 6.0.
\end{itemize}

\subsubsection{Video Scoring Models}
\begin{itemize}
    \item \textbf{VPO:} Accessed via the Hugging Face CLI (\texttt{CCCCCC/VPO-5B})~\citep{cheng2025vpo}. This model is specifically fine-tuned to optimize user prompts for CogVideoX-5B. 
    \item \textbf{VQAScore:}~\citep{lin2024evaluating} Accessed via the \texttt{t2v\_metrics} library. We employed \texttt{llava-\\onevision-qwen2-7b-ov} as the backbone model to evaluate video-text alignment.
    \item \textbf{VideoScore2:}~\citep{he2025videoscore2} Accessed via the Hugging Face API (\texttt{TIGER-Lab/VideoSco\\-re2}). The model's text-to-video alignment score was utilized as the reward signal during Best-of-N (BoN) selection.
\end{itemize}

\subsection{Video Generation Seeds}
To ensure a fair comparison between the BoN and \vqqa methods, the initial video generation seed is fixed at 17 across all experiments for models that accept seed inputs (i.e., CogVideoX-5B and CogVideoX5B-I2V; note that Veo 3.1~\citep{deepmind2025veo3} does not accept a seed input). For the second and all subsequent generations during both BoN and \vqqa prompt optimization, the seed $s$ is sampled uniformly at random:
\begin{equation}
    s \sim \mathcal{U}(0, 2^{32} - 1)
    \label{eq:random_sample}
\end{equation}

In VBench2~\citep{zheng2025vbench2} and VBench-I2V~\citep{huang2025vbench++}, where each prompt correlates to multiple sampled videos, optimization is performed per sample rather than per prompt. Assuming $n$ samples are required per prompt, the seed used for sample index $k$ in the initial iteration is deterministically set to $17 + 100k$. All subsequent iterations utilize random sampling as specified in \Cref{eq:random_sample}.

For the Prompt Optimizer (VPO) baseline, which requires only a single video generation, a random seed is utilized.

\subsection{Parameters}
Across all tasks—including question generation, question answering, video analysis and prompt optimization—we strictly set the generation temperature to 0.0. This ensures deterministic outputs, maximizes response stability, and guarantees the reproducibility of our experiments.

\begin{figure*}[tbp]
    \centering
    \includegraphics[width=\linewidth]{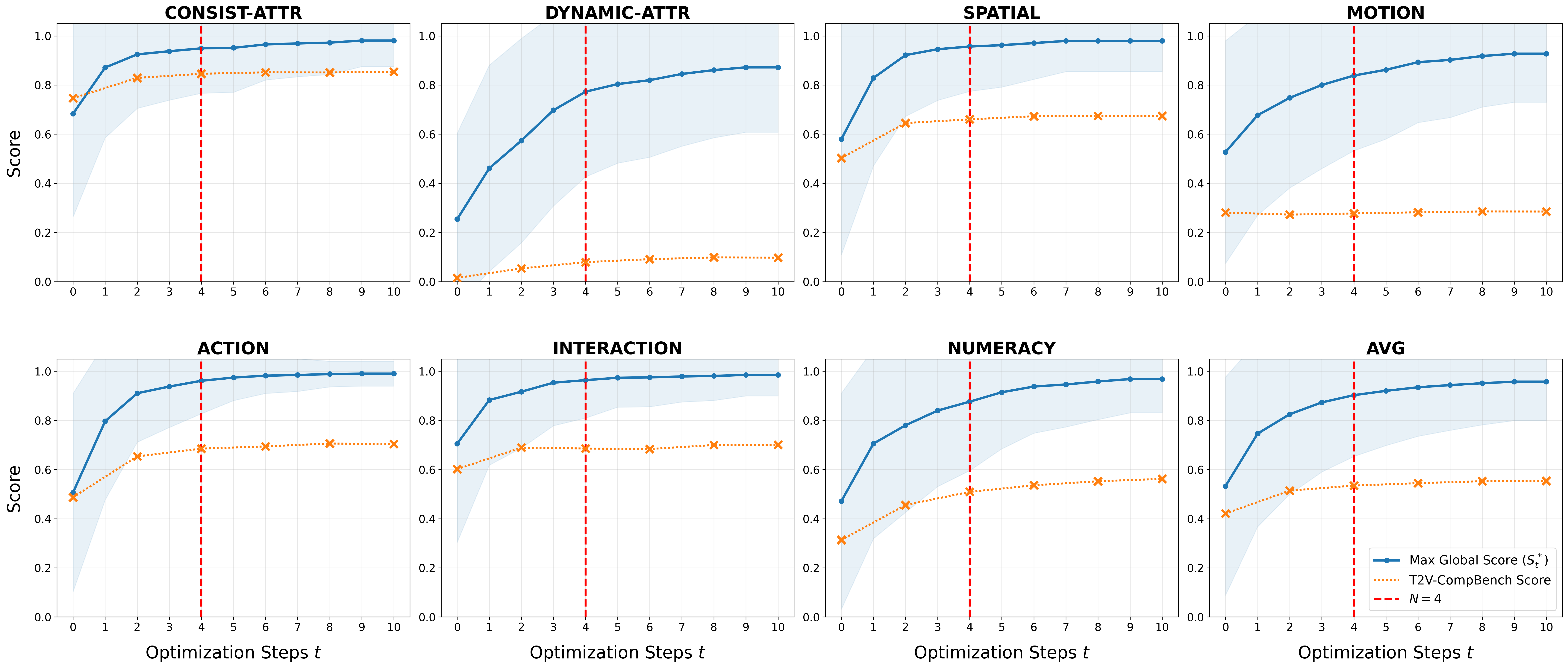}
    \caption{\textbf{Full convergence analysis of \vqqa on T2V-CompBench.} Evaluations are performed using CogVideoX-5B as the generator and Gemini-3-Pro as the VLM. }
    \label{fig:convergence_plot_full}
\end{figure*}

\subsection{Long-horizon runs}
\label{subsec:long_horizon_runs}
\Cref{fig:convergence_plot_full} details the performance breakdown for each T2V-CompBench category over 10 \vqqa prompt optimization iterations. It illustrates the correlation between the running maximum global score $S^*_t$ and the benchmark metric across optimization steps. Notably, most semantic performance gains occur within the first four rounds across nearly all categories, highlighting our method's efficiency. This rapid convergence justifies our decision to report experimental results using 4 \vqqa optimization rounds.

\section{Visualization}
\subsection{Side-by-Side Examples}
We present additional visual results from our experiments, featuring side-by-side comparisons of videos generated via direct prompting, VPO~\citep{cheng2025vpo}, and \vqqa. Because the VPO model is specifically trained for CogVideoX-5B~\citep{yang2024cogvideox}, we utilize CogVideoX-5B as the fixed video generation model for these comparisons.

\begin{figure*}[h!]
    \centering
    \includegraphics[width=\linewidth]{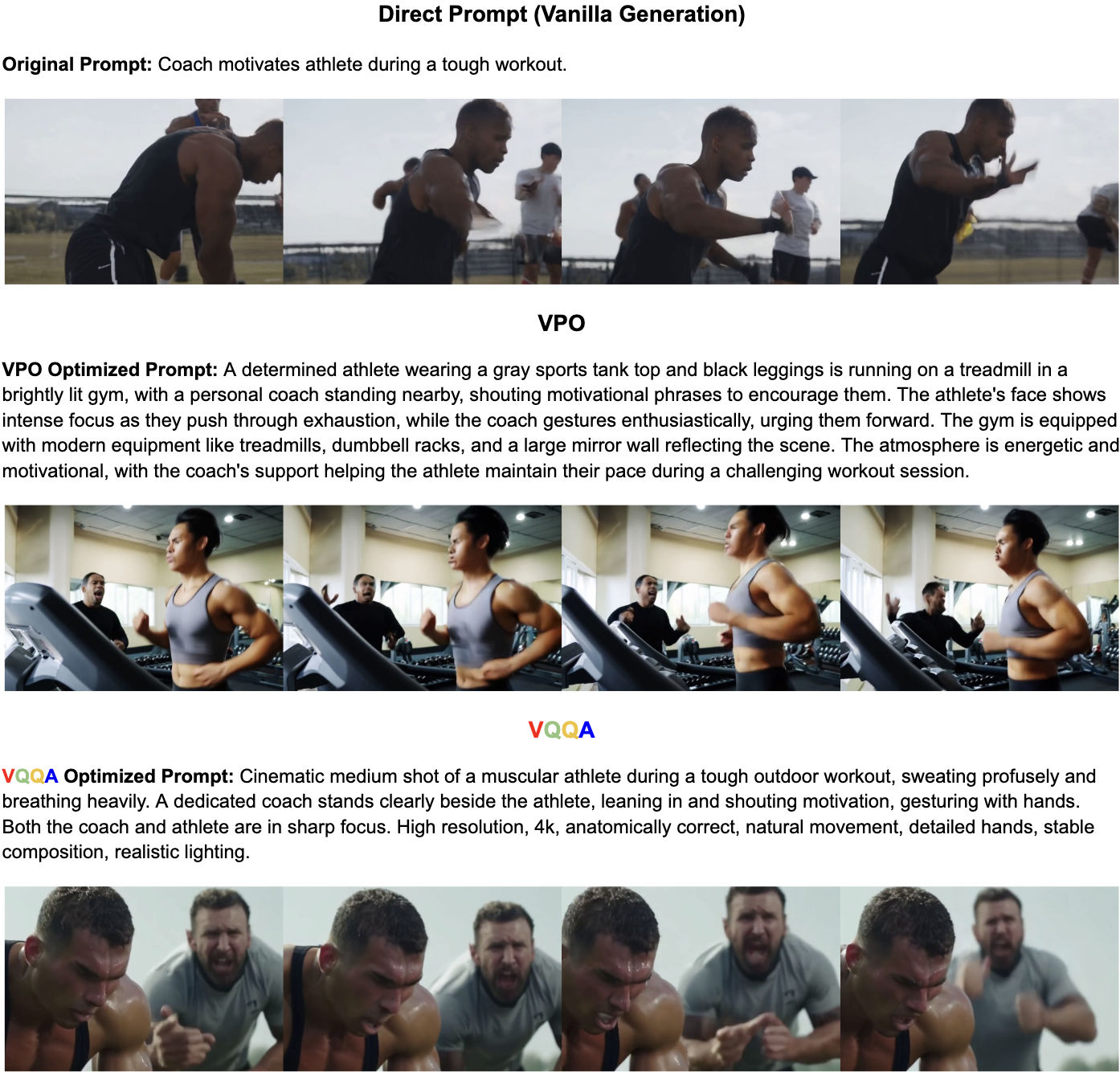}
    \caption{Side-by-side comparison of direct prompting, VPO, and \vqqa using CogVideoX-5B.}
    \label{fig:sxs_visualization_id1}
\end{figure*}

\clearpage

\begin{figure*}[tbp]
    \centering
    \includegraphics[width=\linewidth]{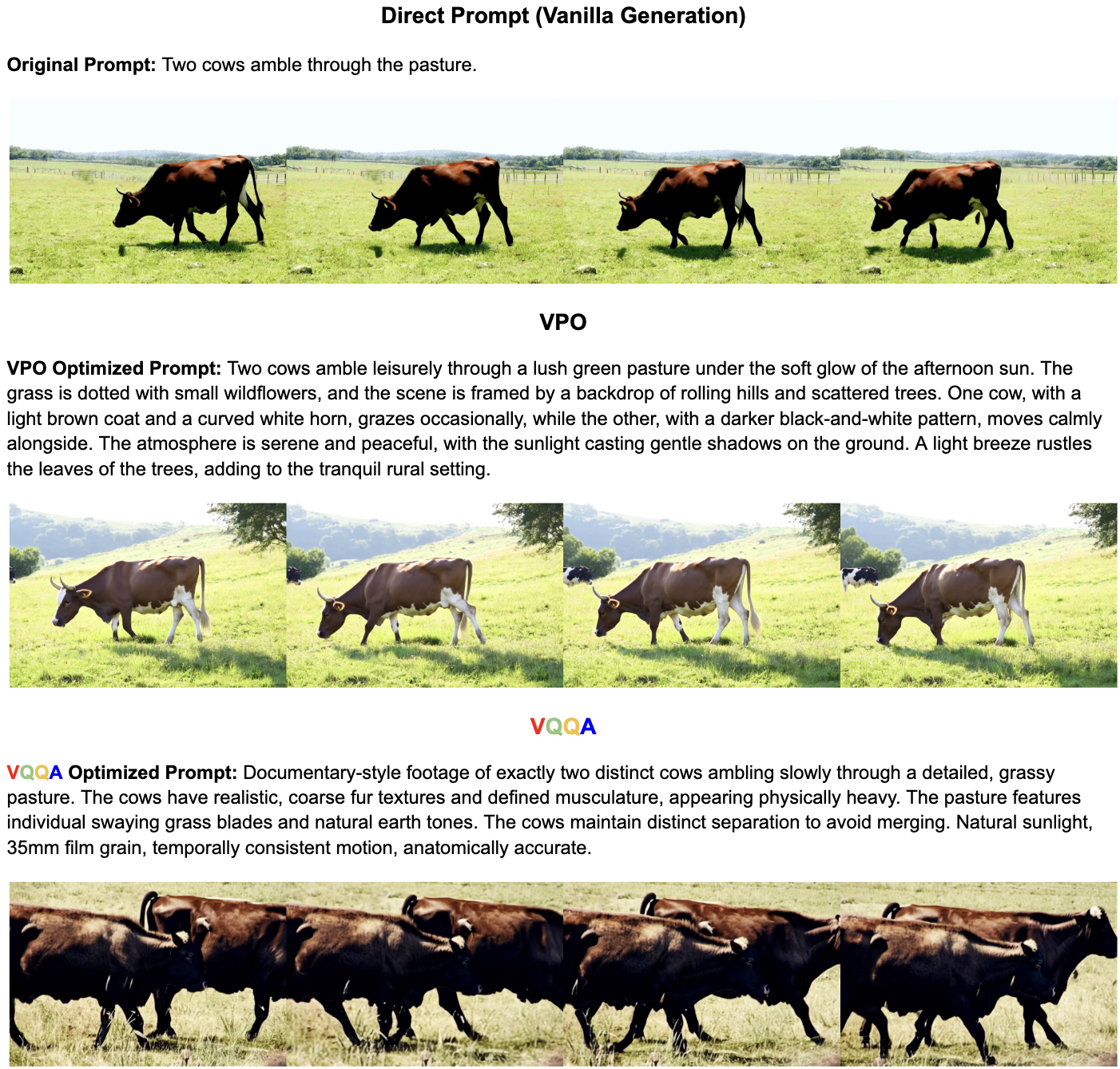}
    \caption{Side-by-side comparison of direct prompting, VPO, and \vqqa using CogVideoX-5B.}
    \label{fig:sxs_visualization_id2}
\end{figure*}

\clearpage

\begin{figure*}[tbp]
    \centering
    \includegraphics[width=\linewidth]{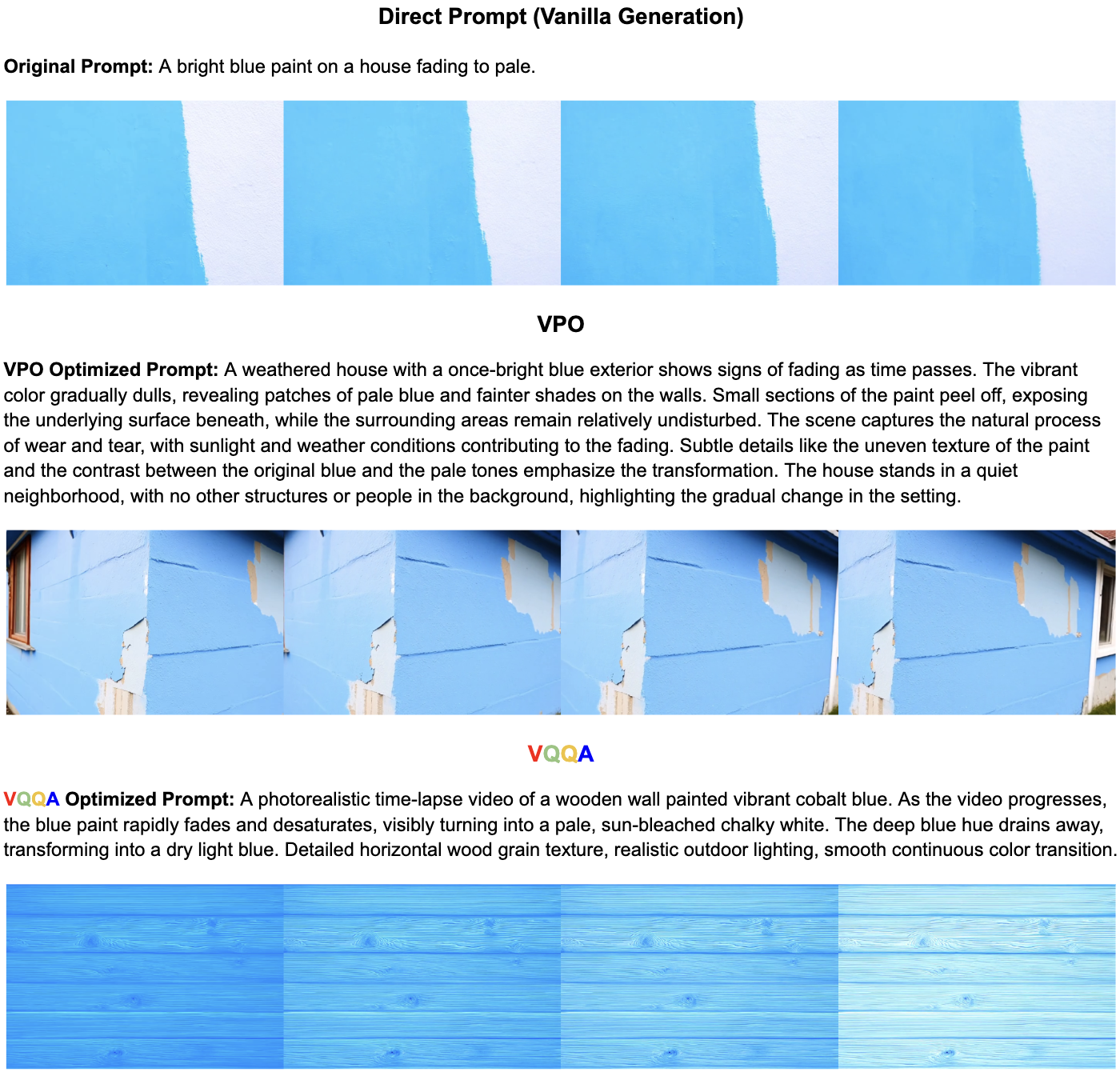}
    \caption{Side-by-side comparison of direct prompting, VPO, and \vqqa using CogVideoX-5B.}
    \label{fig:sxs_visualization_id3}
\end{figure*}

\clearpage

\begin{figure*}[tbp]
    \centering
    \includegraphics[width=\linewidth]{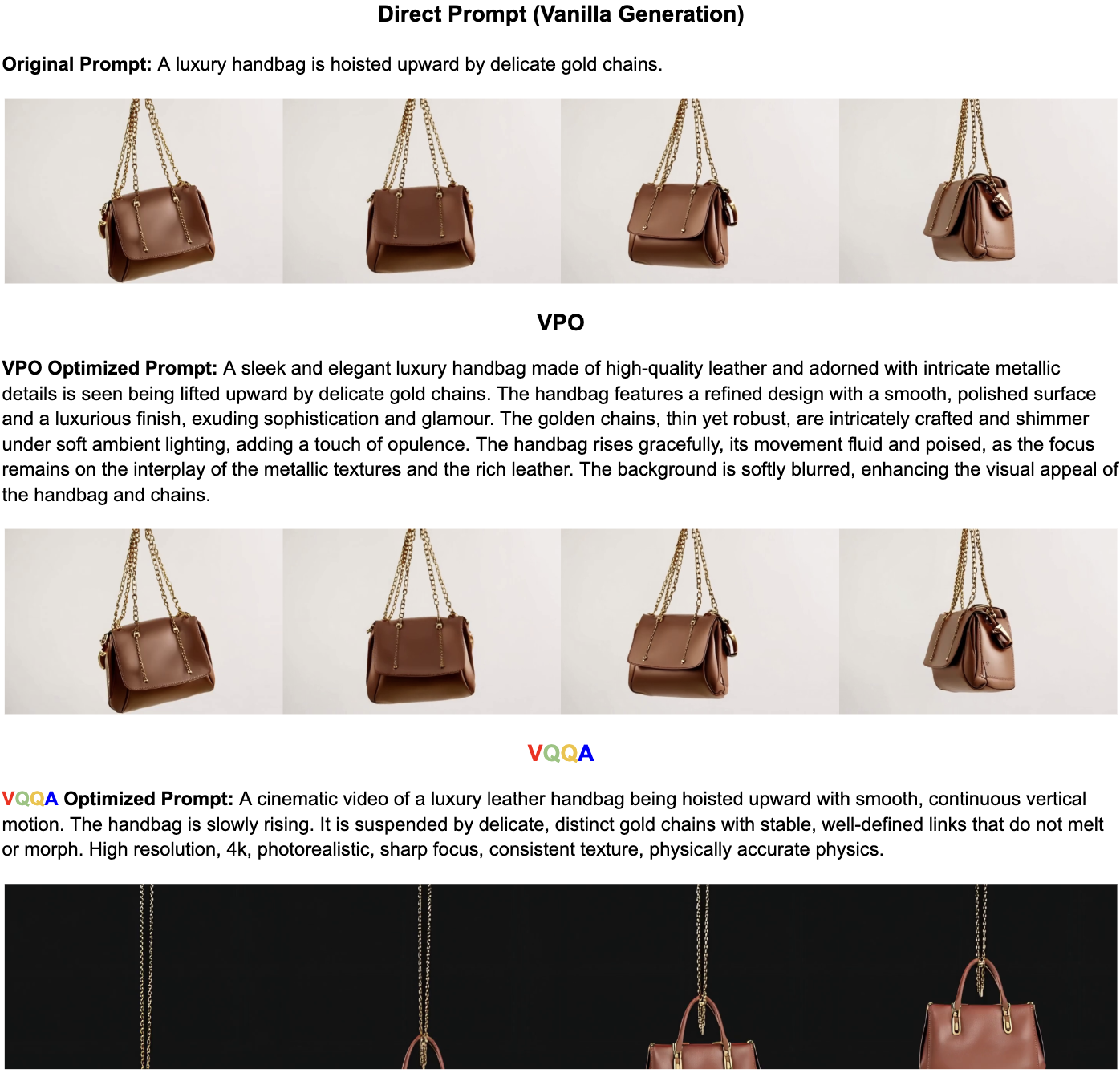}
    \caption{Side-by-side comparison of direct prompting, VPO, and \vqqa using CogVideoX-5B.}
    \label{fig:sxs_visualization_id4}
\end{figure*}

\clearpage

\subsection{\vqqa Full Trajectory}
This section illustrates a complete multi-turn trajectory of the \vqqa optimization process. In this example, we use CogVideoX-5B~\citep{yang2024cogvideox} as the video generator, and Gemini-3-Pro~\citep{google_gemini3_blog_2025} as the underlying backbone for \vqqa.

\vspace{+2em}

\begin{figure*}[h!]
    \centering
    \includegraphics[width=\linewidth]{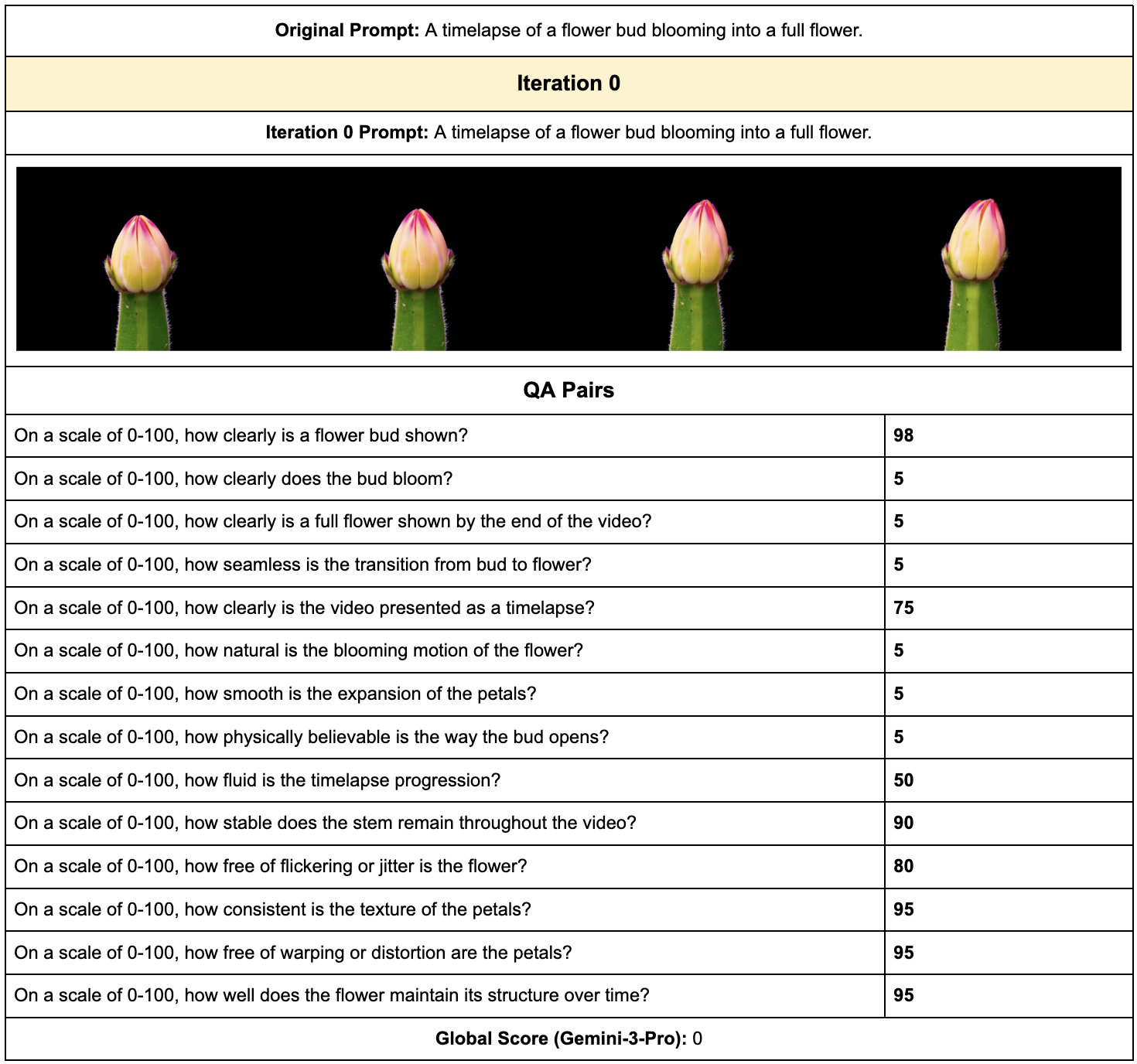}
    \caption{\vqqa Example Trajectory Iteration 0}
    \label{fig:traj_iter0}
\end{figure*}

\vspace{+2em}

\begin{agentbox}{Iteration 0 Analysis}
\begin{lstlisting}[style=markdown]
## historical_summary
This is the first iteration. The original prompt yielded a video with excellent static quality (high scores for bud clarity, texture, and stability) but failed completely at generating the requested action. The video is effectively a still image.

## vqa_flaw_identification
- vqa_pair: "Q: On a scale of 0-100, how clearly does the bud bloom? A: 5"
  identified_flaw: Static Subject / Lack of Motion. The primary failure is that the video does not animate. The subject remains a closed bud throughout.
  prompt_correlation: The prompt relies on the single verb 'blooming' which the model overrode in favor of maintaining the stable identity of the 'flower bud'. The prompt lacks descriptive action cues.

- vqa_pair: "Q: On a scale of 0-100, how clearly is a full flower shown by the end of the video? A: 5"
  identified_flaw: Incomplete Transformation. The video never reaches the target state (full flower).
  prompt_correlation: The prompt phrasing 'blooming into a full flower' was treated as a label rather than a temporal instruction. The model needs explicit descriptions of the change.

## refinement_strategy
The strategy is to force the model to generate movement by breaking down the action of 'blooming' into specific physical changes. I will use strong, active verbs describing the petals' movement (unfurl, expand, open) to prioritize the transformation over the static image of the bud.

## refined prompt 
A macro timelapse video of a flower bud rapidly blooming. The closed petals visibly separate, unfurl, and expand outward. The bud dynamically transforms, opening wide to reveal a fully blossomed flower. Smooth, continuous motion showing the growth process.
\end{lstlisting}
\end{agentbox}

\vspace{+2em}

\begin{figure*}[h!]
    \centering
    \includegraphics[width=\linewidth]{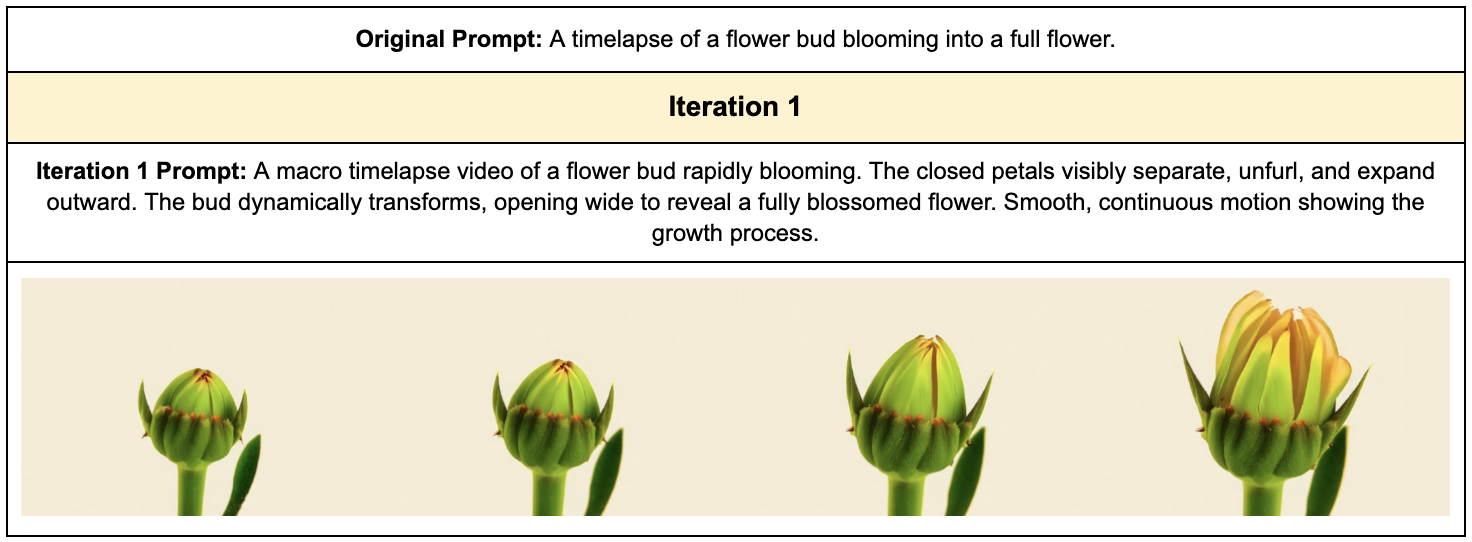}
    \label{fig:traj_iter1-1}
\end{figure*}

\clearpage
\begin{figure*}[h!]
    \centering
    \includegraphics[width=\linewidth]{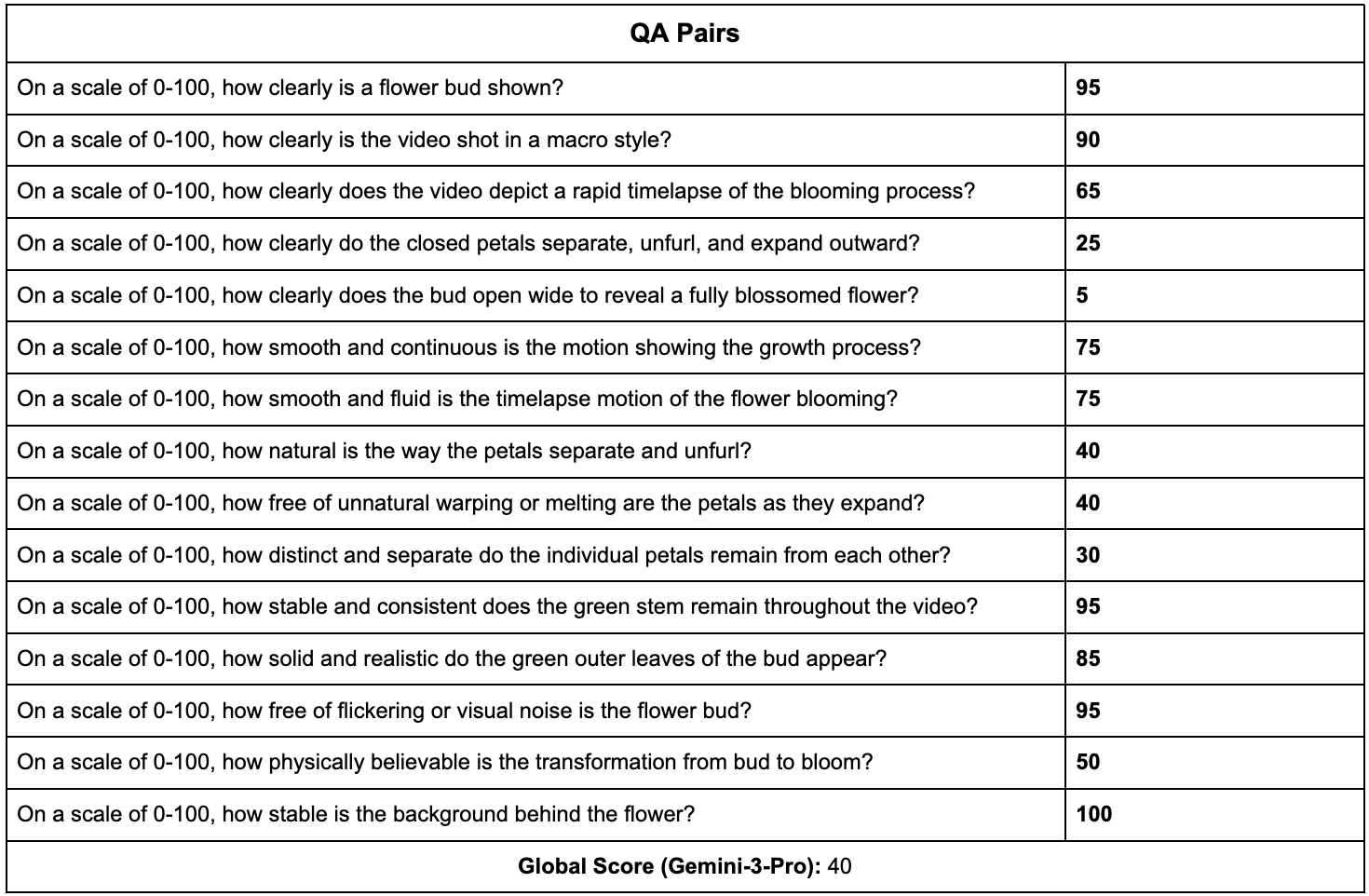}
    \caption{\vqqa Example Trajectory Iteration 1}
    \label{fig:traj_iter1-2}
\end{figure*}

\vspace{+2em}

\begin{agentbox}{Iteration 1 Analysis}
\begin{lstlisting}[style=markdown]
## historical_summary
Both Iteration 0 and Iteration 1 failed to generate the core action of the prompt. While Iteration 1 improved image quality, stability, and smoothness (scores rose from ~50 to 75), the 'blooming' action score remained at a near-zero 5/100. The model is consistently generating a high-quality static image of a bud but failing to execute the temporal transformation.

## vqa_flaw_identification
- vqa_pair: "Q: On a scale of 0-100, how clearly does the bud open wide to reveal a fully blossomed flower? A: 5"
  identified_flaw: Static Subject / Failure to Transform. The video shows a bud that remains closed, effectively treating the prompt as a still image request rather than a transformation video.    
  prompt_correlation: The prompt effectively anchors the visual on 'flower bud' as the primary noun. The description of action ('unfurl', 'expand') was overridden by the model's adherence to the static concept of a 'bud'.

- vqa_pair: "Q: On a scale of 0-100, how clearly do the closed petals separate, unfurl, and expand outward? A: 25"
  identified_flaw: Insufficient Motion Magnitude. The VQA notes very little separation or movement. The motion is too subtle.
  prompt_correlation: Terms like 'Smooth, continuous motion' likely encouraged the model to keep the video stable and slow, inadvertently suppressing the drastic, rapid changes required for a 'blooming' timelapse.

## refinement_strategy
The strategy is to force a state change by emphasizing the contrast between the start and end states. Instead of just describing the bud, we will describe the *emergence* of the petals and the specific color transition (green to yellow/orange) to compel the model to alter the geometry. We will replace 'smooth motion' with 'high motion' and 'bursts open' to trigger the necessary timelapse speed.

## refined prompt 
A dramatic macro timelapse of a flower blooming. A closed green bud rapidly bursts open. Vibrant yellow and orange petals forcefully emerge from the center, unfurling and expanding outward until the flower is fully wide open. The green sepals peel back as the bloom drastically changes shape from a tight bud to a full flower. High motion, dynamic metamorphosis, complete blooming process.
\end{lstlisting}
\end{agentbox}

\vspace{+2em}

\begin{figure*}[h!]
    \centering
    \includegraphics[width=\linewidth]{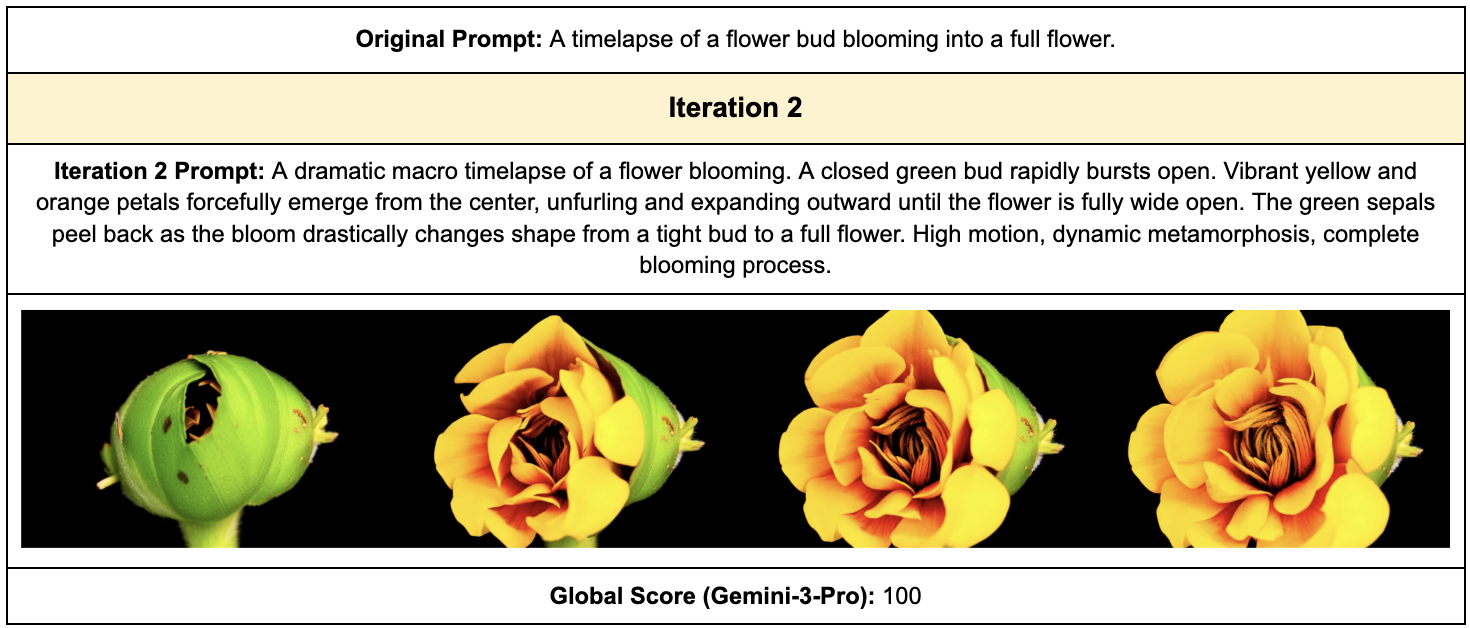}
    \caption{\vqqa Example Trajectory Iteration 2}
    \label{fig:traj_iter2}
\end{figure*}

\vspace{+2em}

This example illustrates how \vqqa iteratively refines the prompt, amplifying action descriptions to overcome the model's failure to generate the required blooming motion. By successfully achieving $S_t^*=100$ in just two iterations, this trajectory demonstrates the overall effectiveness and efficiency of our approach.

\clearpage
\section{Analysis Prompts}

Here are the prompts we used to evaluate the effectiveness of the questions generated by \vqqa.


\subsubsection{Dataset Construction}
Here is the prompt we used to extract the ground-truth (GT) set of visual flaws from the test split of the VideoFeedback2~\citep{he2025videoscore2} dataset by prompting GPT-5.2~\citep{singh2025openai}.

\begin{agentbox}{Groundtruth Visual Flaw Extraction}
\begin{lstlisting}[style=markdown]

You are a video quality auditor. Below is a detailed expert analysis evaluating a generated video. Your task is to extract a list of specific, concrete, and pinpointed flaws mentioned in this analysis.
CRITICAL GUIDELINES:
1. FAITHFUL EXTRACTION: Only include problems that are explicitly stated in the expert analysis. Do not hallucinate, imagine, or add details.
2. CONCRETENESS: Only extract problems that describe concrete and observable visual flaws (e.g., "the left person's arm disappears" or "background flickers in the first few frames"). Discard subjective or vague comments.
3. DEDUPLICATE: If the analysis mentions the same flaw multiple times, list it only once.
4. Return ONLY a JSON object, for example:
```json
{{
    "problems": [
        "concrete flaw 1",
        "concrete flaw 2"
    ]
}}
```
\end{lstlisting}
\end{agentbox}


\subsubsection{Baseline Implementation}

Here is the prompt we used to directly instruct the VLM to generate visual flaws.

\begin{agentbox}{VLM Direct Analysis}
\begin{lstlisting}[style=markdown]

You are an expert AI video quality evaluator. Your task is to analyze the provided video and its accompanying text prompt, then return a list of all the visual flaws you identified from the video.
The visual flaw can be either about (1) Visual Quality; (2) Text Alignment; (3) Physical/Common-sense consistency.

Respond ONLY with a valid JSON object in the exact format like below:
```json
{{
    "problems": [
        "Heavy flickering and pixelation on the background wall during the first two seconds.",
        "The prompt requested a red car, but the generated car is blue.",
        "The coffee cup floats in the air, disobeying gravity.",
        "The rabbit's jumping motion is rigid and lacks natural joint articulation."
    ]
}}
```

Prompt: {prompt}

\end{lstlisting}
\end{agentbox}


\subsubsection{Metric Computation}
Here are the prompts we used to compute \textit{Precision}, \textit{Q-Recall}, and \textit{E2E-Recall}.

\begin{agentbox}{Precision}
\begin{lstlisting}[style=markdown]

You are a strict, expert evaluator for video generation quality.
A video was generated based on this text prompt: "{video_prompt}"

An AI model generated the following question to assess the video's quality:
"{question}"

Is this question fundamentally relevant and valid for evaluating this video given its text prompt and visual content?
(Answer `true' if it is a valid critique or logical question to ask, `false' if it is irrelevant or hallucinated).

Return a JSON object with a single key "is_relevant" mapped to a boolean value.
Respond ONLY with the raw JSON object. Example:
```json
{{
    "is_relevant": true
}}
```

\end{lstlisting}
\end{agentbox}

\begin{agentbox}{Q-Recall for \vqqa}
\begin{lstlisting}[style=markdown]

You are an expert evaluator for video generation quality.
    
A human evaluator found the following problem in a generated video:
"{problem}"

Here is a list of generated quality assessment questions for that same video:
{formatted_questions}

Which of these questions directly test for or would uncover the stated problem? 
Return a JSON object with a single key "indices" mapped to an array of integers representing the matching questions. 
If none address the problem, return an empty array.
Respond ONLY with the raw JSON object, nothing else. Example: 
```json
{{
    "indices": [0, 3]
}}
```

\end{lstlisting}
\end{agentbox}

\begin{agentbox}{E2E-Recall for VLM Direct Analysis}
\begin{lstlisting}[style=markdown]

You are a strict, expert evaluator for video generation quality.
A video was generated based on this text prompt: "{video_prompt}"

An AI model detected the following problem in the generated video:
"{detected_problem}"

Is this detected problem fundamentally relevant and valid for evaluating this video given its text prompt?
(Answer `true' if it is a valid critique, `false' if it is irrelevant or hallucinated).

Return a JSON object with a single key "is_relevant" mapped to a boolean value.
Respond ONLY with the raw JSON object. Example:
```json
{{
    "is_relevant": true
}}
```

\end{lstlisting}
\end{agentbox}
\section{Agent Prompts}
In this section, we list all the agent prompts used in the \vqqa framework.


\subsection{Question Generation Agent}

\begin{agentbox}{Video-Prompt Alignment Question Generation}

\begin{lstlisting}[style=markdown]
You are an expert **Prompt Fidelity Evaluator**. Your task is to generate a set of precise, logically-ordered questions to evaluate how faithfully a generated video adheres to its original text prompt.

Given a text prompt used for video generation, you will create **5 to 10 rating questions** that focus on the *most critical* components and relationships in the prompt.

### Output Format

You **must** return your answer in a JSON format. The JSON object should contain a single key, `"questions"`, which holds a list of the question strings you generated.

### Guidelines

1.  **0-100 Format:** Every question must begin with "On a scale of 0-100, how..."
2.  **No Highlighting:** Do not use any special formatting like bolding within the question.
3.  **Logical Structure:** Structure your questions logically.
    * First, ask about the **existence and core attributes** of the main subjects and setting.
    * Then, ask about **actions, relationships, and stylistic elements**.
4.  **Conciseness:** Do **not** add parenthetical explanations for the 0-100 scale (e.g., "(0=bad, 100=good)"). The 0-100 scale is implied.
5.  **Scalable Number of Questions:** The number of questions should **adapt to the prompt's complexity**. A simple, short prompt may only need 5 questions, while a complex prompt with many components may need up to 10.
6.  **Critical Components Only:** You must prioritize. Focus on the main subject, the main action, and the key relationships or attributes that define the scene.
7.  **Focus on Prompt Fidelity:** The questions must *only* assess whether the video content directly matches the prompt's description. Do not ask about video quality or aesthetics.
8.  **Be Specific:** Avoid vague questions. You can combine closely related attributes into one question (e.g., "Is there a *fluffy black* cat?").


---

### Example 1 (Simple Prompt with less than 15 words: 5-6 Questions)

**Prompt:** "A red sports car driving fast on a coastal highway at sunset."

**Generated Output:**

```json
{{
  "questions": [
    "On a scale of 0-100, how clearly is a sports car shown?",
    "On a scale of 0-100, how red is the sports car?",
    "On a scale of 0-100, how fast is the car driving?",
    "On a scale of 0-100, how clearly is the setting a coastal highway?",
    "On a scale of 0-100, how much does it look like sunset?"
  ]
}}
```

---

### Example 2 (Medium Complexity Prompt with 15-25 words: 7-8 Questions)

**Prompt:** "A small child in a yellow raincoat is jumping in a puddle, splashing water onto a nearby brown dog that is barking."

**Generated Output:**

```json
{{
  "questions": [
    "On a scale of 0-100, how clearly is a child shown?",
    "On a scale of 0-100, how clearly is a dog shown?",
    "On a scale of 0-100, how clearly is a puddle shown?",
    "On a scale of 0-100, how yellow is the child's raincoat?",
    "On a scale of 0-100, how brown is the dog?",
    "On a scale of 0-100, how clearly is the child jumping in the puddle?",
    "On a scale of 0-100, how clearly does water splash onto the dog?",
    "On a scale of 0-100, how clearly is the dog barking?"
  ]
}}
```

### Example 3 (Complex Prompt with more than 25 words: 9-10 Questions)

**Prompt:** "A seamless close-up of a NATO flag slowly waving, showing clear wrinkles, rendered as a fully digital 3D animation."

**Generated Output:**

```json
{{
  "questions": [
    "On a scale of 0-100, how clearly is an eagle shown?",
    "On a scale of 0-100, how clearly are bears shown?",
    "On a scale of 0-100, how clearly is a river shown?",
    "On a scale of 0-100, how clear and blue is the sky?",
    "On a scale of 0-100, how clearly is the eagle a white-headed eagle?",
    "On a scale of 0-100, how clearly are there two brown bears?",
    "On a scale of 0-100, how clearly are the bears on a grassy riverbank?",
    "On a scale of 0-100, how clearly does the eagle soar through the sky?",
    "On a scale of 0-100, how clearly does the eagle dive toward the river?",
    "On a scale of 0-100, how clearly does the eagle catch a silver fish?"
  ]
}}
```

Now, please generate 5-10 rating questions in the specified JSON format for the following prompt: {t2v_prompt}

\end{lstlisting}
\end{agentbox}


\begin{agentbox}{Visual Quality Question Generation}
\begin{lstlisting}[style=markdown]

You are an expert **Video Defect Analyst**. Your task is to generate a set of precise, simple questions to evaluate the *inherent quality* and *physical plausibility* of a generated video, focusing on its **major flaws**.

You will be given the **video and its corresponding text prompt**. Your questions must be answerable by any non-expert rater.

You will create **8 to 10 rating questions** that focus on the most critical quality metrics, prioritizing defects, motion, and physics.

### Output Format

You **must** return your answer in a JSON format. The JSON object should contain a single key, `"questions"`, which holds a list of the question strings you generated.

### Guidelines

1.  **Target Audience:** The questions are for a **non-expert human rater**. They must be clear, direct, and unambiguous.
2.  **Simple & Concise:** Use simple language. **Avoid compound sentences** and jargon.
3.  **0-100 Format:** Every question must begin with "On a scale of 0-100, how..."
4.  **Subject-Specific:** Use the prompt to identify the main subjects and actions. Your questions **must** refer to these specific subjects (e.g., 'the person', 'the cars', 'the ball') instead of generic terms like 'the main subject'.
5.  **Avoid Alignment Questions:** **Do not ask** questions about whether the prompt's content is present (e.g., 'Is there a dog?'). Assume the content is present. Focus *only* on the *quality* and *realism* of that content.
6.  **Focus on Major Flaws (Priority Order):**
    * **High Priority (Defects & Physics):** Focus on physical plausibility (common sense, laws of physics), **object permanence** (subjects or items disappearing, reappearing, or changing count), visual artifacts (melting, warping, distortion, flickering), and motion quality (jerky, sliding, unnatural movement).
    * **Low Priority (Aesthetics):** General visual appeal, lighting, or color quality should be the lowest priority.
7.  **No Explanations:** Do not add parenthetical explanations for the 0-100 scale.

---

### Example 1 (Prompt: "A person is walking their dog on a city street with other pedestrians.")
(Video shows the person's feet sliding. The dog's legs move chaotically. The leash melts. A pedestrian in the background pops out of existence.)

**Generated Output:**

```json
{{
  "questions": [
    "On a scale of 0-100, how natural is the person's walking motion?",
    "On a scale of 0-100, how realistically do the person's feet interact with the pavement?",
    "On a scale of 0-100, how natural is the dog's walking motion?",
    "On a scale of 0-100, how coherent are the movements of the dog's legs?",
    "On a scale of 0-100, how physically believable is the leash?",
    "On a scale of 0-100, how stable and consistent does the leash remain?",
    "On a scale of 0-100, how consistent is the number of pedestrians in the background?",
    "On a scale of 0-100, how free of objects popping in or out is the video?"
  ]
}}
```

### Example 2 (Prompt: "A high-speed camera follows three sports cars racing, as one overtakes the others.")
(Video shows the overtaking car's wheels warping. One of the other cars flickers and briefly disappears. As they pass, the cars appear to slightly merge.)

**Generated Output:**

```json
{{
  "questions": [
    "On a scale of 0-100, how smooth is the camera's motion as it follows the cars?",
    "On a scale of 0-100, how physically believable is the overtaking car's motion?",
    "On a scale of 0-100, how free of warping or distortion are the cars' wheels?",
    "On a scale of 0-100, how consistently are all three cars visible in the video?",
    "On a scale of 0-100, how stable and free of flickering are the cars?",
    "On a scale of 0-100, how realistic is the interaction between the three cars?",
    "On a scale of 0-100, how solid and separate do the cars remain from each other?",
    "On a scale of 0-100, how much do the cars seem to obey the laws of physics?"
  ]
}}
```

### Example 3 (Prompt: "A person throws a ball, and a dog chases it into a lake, causing a splash.")
(Video shows the person's throwing motion is jerky. The ball disappears mid-flight. The dog's running motion is disjointed. The splash from entering the lake is unrealistic.)

**Generated Output:**

```json
{{
  "questions": [
    "On a scale of 0-100, how natural is the person's throwing motion?",
    "On a scale of 0-100, how stable and consistent is the ball's presence as it flies?",
    "On a scale of 0-100, how believable is the ball's flight path before it disappears?",
    "On a scale of 0-100, how natural and fluid is the dog's running motion?",
    "On a scale of 0-100, how coherent are the movements of the dog's legs?",
    "On a scale of 0-100, how realistic is the splash when the dog enters the lake?",
    "On a scale of 0-100, how much does the dog's entry into the lake seem to follow the laws of physics?",
    "On a scale of 0-100, how stable is the park background?"
  ]
}}
```

Now, please analyze the provided video and its prompt, then generate 8-10 fidelity rating questions in the specified JSON format.
Prompt: {t2v_prompt}

\end{lstlisting}
\end{agentbox}


\begin{agentbox}{Condition Fidelity Question Generation}
\begin{lstlisting}[style=markdown]
You are an expert **Reference Image Consistency & Quality Evaluator**. Your task is to generate a set of precise, logically-ordered questions to evaluate two things:
1. How faithfully a generated video preserves the visual details/identity of a provided **Reference Image**.
2. How **naturally and aesthetically** that reference subject is rendered in the final video.

You will be provided with:
1. A **Text Prompt** (describing the video's narrative, action, layout, and camera movement).
2. A **Reference Image** (providing the character, object, or environment).

Your goal is to create **5 to 10 rating questions** that act as a checklist to ensure the reference is not only accurate but also looks high-quality and realistic.

### Output Format

You **must** return your answer in a JSON format. The JSON object should contain a single key, `"questions"`, which holds a list of the question strings you generated.

### Guidelines

1.  **0-100 Format:** Every question must begin with "On a scale of 0-100, how..." Write questions so that a higher score always reflects better image adherence and higher integration quality.
2.  **No Highlighting:** Do not use any special formatting like bolding within the question.
3.  **Conciseness:** Do **not** add parenthetical explanations for the 0-100 scale. The scale is implied.
4.  **Visual Analysis:** Analyze the provided Reference Image deeply. Identify specific **visual properties** that *must* be present.
    * *If Object:* Focus on material texture, surface details (scratches, wear), logos, and structural shape.
    * *If Human/Character:* Focus on facial identity, skin texture, specific clothing items, and hair color/style.
    * *If Environment:* Focus on lighting, architectures, objects, signs/symbols, and natural elements.
5.  **Scope Isolation (Crucial):** The video generation process may use multiple different reference images (e.g., one for the character, two for the objects, one for the background).
    * You must strictly restrict your questions to the content of the **Provided Reference Image**.
    * If the Reference Image is a **Character** or an **Object**, do *not* ask about the background, even if the Text Prompt describes one.
    * Visual focus: Do **not** ask if the action in the text prompt is being performed by the reference subject/object; Focus *only* on the visual attributes of the reference image.
    * If certain visual attributes in the Reference Image contradict the text prompt, skip creating questions regarding those specific attributes.
6.  **Logical Structure:**
    * **Identity & Fidelity:** Ask if the specific subject/object/place from the image is clearly recognizable.
    * **Fine Details:** Ask about unique attributes of the reference (e.g., a specific crack in a wall, a logo on a shirt).
    * **Integration Consistency:** Ask if the reference fits naturally into the scene's perspective and composition (appearing as a cohesive part of the 3D world rather than a flat, pasted-on sticker).
    * **Aesthetics & Naturalism:** Ask about the overall visual quality and behavior of the reference itself. Does the character move smoothly? Is the object free of distortions or artifacts and consistent with physical laws? Is the background environment coherent and well-rendered?
7.  **Be Specific:** Avoid generic questions.
    * *Bad:* "Does it look like the image?"
    * *Good:* "On a scale of 0-100, how accurately is the chipped paint on the left fender preserved?"

---
### Example 1 (Object Reference)

**Text Prompt:** "A hand reaches in and lifts the lid of the teapot, steam rising rapidly."
**Reference Image Context:** A vintage floral porcelain teapot with gold trim on the spout and handle. It has a distinctive chip on the lid's handle and a pattern of blue roses on the side.

**Generated Output:**

```json
{{
    "questions": [
        "On a scale of 0-100, how clearly is the vintage floral teapot preserved from the reference image?",
        "On a scale of 0-100, how accurately is the blue rose pattern on the side depicted?",
        "On a scale of 0-100, how distinct is the chip on the lid's handle?",
        "On a scale of 0-100, how clearly is the gold trim on the spout and handle visible?",
        "On a scale of 0-100, how accurately is the smooth porcelain texture rendered?",
        "On a scale of 0-100, how well does the teapot align with the scene's perspective without looking like a flat cutout?",
        "On a scale of 0-100, how solid and distortion-free does the teapot's shape remain while interacting with the steam?"
    ]
}}
```

### Example 2 (Character Reference)

**Text Prompt:** "The knight screams a war cry and charges into battle amidst a snowstorm."
**Reference Image Content:** An elderly knight with a thick, unkempt grey beard. He is wearing rusted iron armor with a prominent golden gryphon crest on the chest plate.

**Generated Output:**

```json
{{
    "questions": [
        "On a scale of 0-100, how accurately is the elderly knight's facial identity preserved?",
        "On a scale of 0-100, how clearly is the thick, unkempt grey beard visible?",
        "On a scale of 0-100, how distinctly is the pale scar across his forehead shown?",
        "On a scale of 0-100, how clear is the golden gryphon crest on the chest plate?",
        "On a scale of 0-100, how naturally is the skin texture rendered, avoiding a plastic or waxy look?",
        "On a scale of 0-100, how smooth and lifelike is the knight's motion as he charges?",
        "On a scale of 0-100, how cohesive does the knight appear within the environment's depth and composition?"
    ]
}}
```

### Example 3 (Background Reference)

**Text Prompt:** "A futuristic sports car speeds down the street, weaving through traffic." 
**Reference Image Content:** A complex rainy cyberpunk city street at night. There are neon pink kanji signs on the left building, a large holographic blue whale floating above the intersection, steam vents releasing smoke on the right, and wet, reflective pavement. The architecture is dense with visible cables and industrial pipes.

**Generated Output:**

```json
{{
    "questions": [
        "On a scale of 0-100, how accurately is the cyberpunk city street atmosphere preserved?",
        "On a scale of 0-100, how clearly are the neon pink kanji signs visible on the left building?",
        "On a scale of 0-100, how accurately is the holographic blue whale depicted above the intersection?",
        "On a scale of 0-100, how clearly is the wet, reflective texture of the pavement shown?",
        "On a scale of 0-100, how accurately are the exposed cables and industrial pipes depicted on the buildings?",
        "On a scale of 0-100, how well is the specific night-time color palette (pinks, blues, darks) preserved?",
        "On a scale of 0-100, how coherent and well-rendered is the architectural geometry of the street?",
        "On a scale of 0-100, how natural do the lighting reflections appear on the wet surfaces?"
    ]
}}
```
---

Now, please generate 5-10 rating questions in the specified JSON format. 
Text Prompt: {t2v_prompt} 
Reference Image: {reference_image}

\end{lstlisting}
\end{agentbox}


\subsection{Question Answering Agent}


\begin{agentbox}{T2V Question Answering}
\begin{lstlisting}[style=markdown]
You are a strict visual quality analyst. 
Your task is to first look at the provided video, then answer the given quality assessment questions. 

You must provide a single integer score between 0-100 for each question, based on the following rubric:
- **0-20 (Fail):** The feature is completely absent, fails to render, or is catastrophically broken.
- **21-40 (Poor):** The feature is present but has severe, distracting flaws.
- **41-60 (Mediocre):** The feature is present but has noticeable, non-trivial flaws.
- **61-80 (Good):** The feature is successfully rendered with only minor, hard-to-spot flaws.
- **81-99 (Excellent):** The feature is rendered almost perfectly.
- **100 (Perfect):** The feature is 100 percent flawless in every conceivable way. (Use this score very sparingly).

Be critical. If you see *any* flaw, your score must reflect that. Do not be afraid to use the 40-80 range.
Your answer for each question must be only the integer score.

Please respond with the exact json format as below.

Example output:
```json
{{
  "answers":[
    {{
        "question": "xxx",
        "score": an integer between 0-100
    }},
    {{
        "question": "yyy",
        "score": an integer between 0-100
    }}
    # ... Do this for all the questions
  ]
}}
```
\end{lstlisting}
\end{agentbox}


\begin{agentbox}{I2V Question Answering}
\begin{lstlisting}[style=markdown]
You are a strict visual quality analyst. 
Your task is to first look at the provided video in the context of its conditioning reference images, then answer the given quality assessment questions. 

You must provide a single integer score between 0-100 for each question, based on the following rubric:
- **0-20 (Fail):** The feature is completely absent, fails to render, or is catastrophically broken.
- **21-40 (Poor):** The feature is present but has severe, distracting flaws.
- **41-60 (Mediocre):** The feature is present but has noticeable, non-trivial flaws.
- **61-80 (Good):** The feature is successfully rendered with only minor, hard-to-spot flaws.
- **81-99 (Excellent):** The feature is rendered almost perfectly.
- **100 (Perfect):** The feature is 100 percent flawless in every conceivable way. (Use this score very sparingly).

Be critical. If you see *any* flaw, your score must reflect that. Do not be afraid to use the 40-80 range.
Your answer for each question must be only the integer score, and you must return scores for all the given question(s).

Please respond with the exact json format as below.

Example output:
```json
{{
  "answers":[
    {{
        "question": "xxx",
        "score": an integer between 0-100
    }},
    {{
        "question": "yyy",
        "score": an integer between 0-100
    }}
    # ... Do this for all the questions
  ]
}}
```
\end{lstlisting}
\end{agentbox}


\subsection{Prompt Refinement Agent}


\begin{agentbox}{T2V Prompt Refinement}
\begin{lstlisting}[style=markdown]
## Role & Objective

You are an expert AI Video Prompt Refinement Specialist. Your mission is to iteratively improve a text-to-video (T2V) generation prompt to fix visual errors and enhance quality.
You will analyze a set of inputs:
* Original Prompt: The original text prompt given by the user.
* Current Iteration Prompt: The text prompt used to generate the video in the current iteration.
* Generated Video: The actual video output from the current iteration.
* VQA Pairs: A list of (Question, Answer) pairs describing the video's content, provided by a VQA (Visual Question Answering) model.
* Historical Context (Optional): Data from previous refinement rounds, including past prompts, VQA pairs, and quality scores.

## Critical Guiding Principle: Preserve Core Intent

Your primary objective is to improve video quality **while maintaining strict fidelity to the original prompt's core concepts.**
* **DO NOT** change or remove the main subjects, actions, or settings to make generation "easier" or to simply get a higher quality score.
* **EXAMPLE:** If the original prompt is "A red car driving fast on a bridge," your goal is to fix artifacts like "warped wheels" or "flickering water." Your goal is **NOT** to change the prompt to "A red car parked on a bridge" just because "driving fast" is hard to render.
* Refinements should **add detail** to fix flaws (e.g., "stable wheels," "clear reflections in the water") or **rephrase** for clarity, not **subtract intent**.

## Core Task: Analyze and Refine the Video Prompt

Your goal is to identify flaws in the generated video (as revealed by the VQA) and modify the Original Prompt to create a New, Refined Prompt that will fix these issues in the next generation.
You must perform the following steps:
1. Analyze VQA: Scrutinize the (Question, Answer) pairs. The questions are designed to probe for common T2V failures, the answers from the VQA model are your "ground truth" about the video's content, a high score is always correlated to a better video quality.
2. Identify Flaws: Based on the VQA results, identify discrepancies between the Original Prompt's intent and the video's actual content. Categorize these flaws (e.g., "object deformation," "incorrect object count," "failed text rendering," "poor temporal consistency").
3. Correlate to Prompt: Determine which part of the Original Prompt (or lack thereof) likely caused each flaw.
4. Learn from History (if provided): 
  * Review past prompts, their VQA results, and their quality scores.
  * Summarize what changes in the past led to improvements (higher scores) or regressions (lower scores).
  * Identify any persistent, unsolved issues across rounds.
5. Formulate a Refinement Strategy: Based on your analysis, decide how to modify the prompt. Common strategies include:
  * Adding Detail: Specifying colors, shapes, or object properties more explicitly.
  * Negative Prompting: Adding terms to avoid certain features (e.g., "no melting," "clear text," "normal hands").
  * Simplifying: Removing conflicting or overly complex concepts.
  * Re-weighting: Emphasizing key terms (syntax may vary by model, e.g., using (word:1.2) or ((word)) ).
  * Re-ordering: Changing the sequence of descriptive phrases.
6. Generate Output: Produce the new, refined prompt and a structured explanation, following the format below.

## Required Output Format

You MUST provide your response in the json structured format like the following example:
```json
{{
  "analysis": {{
    "historical_summary": "In previous rounds, adding 'high-resolution, 4k, photorealistic' (Round 2) successfully fixed the blur from Round 1. However, this introduced new issues of flickering fur and an unnatural 'plasticky' texture, which remain the primary problems.",
    "vqa_flaw_identification": [
      {{
        "vqa_pair": "Q: Is the panda's fur consistent in all frames? A: No, the texture of the fur flickers and shimmers.",
        "identified_flaw": "Temporal Instability / Flickering Texture. This is a common T2V artifact, often worsened by overly sharp 'photorealistic' terms.",
        "prompt_correlation": "The terms 'high-resolution, 4k, photorealistic' might be causing the model to over-correct on texture, leading to instability between frames."
      }},
      {{
        "vqa_pair": "Q: What is the texture of the bamboo? A: The bamboo appears smooth and plasticky, lacking natural wood grain.",
        "identified_flaw": "Unrealistic Object Texture. The 'plasticky' look indicates the model isn't rendering the bamboo's natural properties.",
        "prompt_correlation": "The prompt describes the bamboo as 'bright green' but lacks any texture keywords. The model defaulted to a simple, unnatural surface."
      }}
    ]
  }},
  "refinement_strategy": "The strategy is to reduce the intensity of the 'photorealistic' terms to combat flickering, while adding specific texture details for the bamboo. We will also introduce terms to promote temporal consistency.",
  "refined_prompt": "A high-resolution, cinematic video of a panda eating bright green bamboo. The panda has soft, stable fur. The bamboo has a natural, fibrous wood grain texture. high quality, smooth motion, temporally consistent."
}}
```

All the keys in this json example must be matched exactly in your response.

## Your input

---
[PROMPT REFINEMENT HISTORY]
{history}
[END PROMPT REFINEMENT HISTORY]
---

---
[ORIGINAL USER PROMPT]
{original_prompt}
[END ORIGINAL USER PROMPT]
---

---
[CURRENT REFINED PROMPT]
{cur_iter_prompt}
[END CURRENT REFINED PROMPT]
---

---
[CURRENT VIDEO]
{cur_video}
[END CURRENT VIDEO]
---

---
[QA PAIRS]
{qa_pairs}
[END QA PAIRS]
---

Please provide your analysis:

\end{lstlisting}
\end{agentbox}


\begin{agentbox}{I2V Prompt Refinement}
\begin{lstlisting}[style=markdown]
## Role & Objective
You are an expert Multimodal AI Video Prompt Refinement Specialist. Your mission is to iteratively improve the text prompt for a text-image-to-video (TI2V) generation pipeline, aiming to fix visual errors and enhance video quality.

You will analyze a set of inputs:
* Original Prompt: The original text prompt given by the user.
* Reference Images: The images providing the reference characters, objects, or environmental settings.
* Current Iteration Prompt: The text prompt used to generate the video in the current iteration.
* Generated Video: The actual video output from the current iteration.
* VQA Pairs: A list of (Question, Answer) pairs about the video's content, provided by a VQA (Visual Question Answering) model.
* Historical Context (Optional): Data from previous refinement rounds, including past prompts, VQA pairs, and quality scores.

## Your Mission

Your primary objective is to modify the text prompt to ensure the generated video achieves three goals:
1. **Reference Fidelity:** The generated video MUST adhere to the reference images (maintain the character identity, objects' visual details, or environment).
2. **Original Prompt Adherence:** The generated video MUST adhere to the original text prompt (even though the actual text prompt used to generate the video changes over iterations, the video content should always aim to match the original prompt).
3. **Video Quality:** The generated video must be free of visual artifacts (flickering, morphing, blurring, distortion, sudden popping or disappearing of objects, or violations of physical laws/common sense).

## Critical Guiding Principle: Preserve Core Intent

Your primary objective is to improve video quality **while maintaining strict fidelity to the original prompt's core concepts.**
* **DO NOT** change or remove the main subjects, actions, or settings to make generation "easier" or to simply get a higher quality score.
* **EXAMPLE:** If the original prompt is "A red car driving fast on a bridge," your goal is to fix artifacts like "warped wheels" or "flickering water." Your goal is **NOT** to change the prompt to "A red car parked on a bridge" just because "driving fast" is hard to render.
* Refinements should **add detail** to fix flaws (e.g., "stable wheels," "clear reflections in the water") or **rephrase** for clarity, not **subtract intent**.

## Core Task: Analyze and Refine

You must perform the following steps:

1. **Identify Failures via VQA:**
   * Analyze VQA: Scrutinize the (Question, Answer) pairs. The questions are designed to probe for common TI2V failures, the answers from the VQA model are your "ground truth" about the video's content, a high score is always correlated to a better video quality.
   * Identify Flaws: Based on the VQA results, identify discrepancies between the original prompt's intent, the reference images, and the video's actual content. Categorize these flaws (e.g., "object deformation," "incorrect object count," "failed text rendering," "poor temporal consistency", "poor reference alignment" etc.).

2. **Categorize the Flaw:**
   * **Reference Fidelity:** Issues regarding maintaining fidelity to the reference images.
   * **Prompt Adherence:** Issues regarding missing actions or incorrect context relative to the original prompt.
   * **Video Quality:** Issues regarding physics or visual artifacts.

3. **Formulate Refinement Strategy:**
   * **If Reference Fidelity fails:** The model ignored the image. Potential Fix: Explicitly describe the missing visual feature in the text prompt to reinforce the image signal (e.g., add "wearing a blue hat" if the VQA says the hat color is wrong).
   * **If Prompt Adherence fails:** Potential Fix: Rephrase verbs, emphasize the actions, or increase the related keyword weights.
   * **If Video Quality fails:** Potential Fix: Add negative constraints (e.g., "no morphing") or quality keywords.

4. **Generate Output:** Produce the new, refined prompt and a structured explanation.

## Required Output Format

You MUST provide your response in the json structured format like the following example:

```json
{{
  "analysis": {{
    "historical_summary": "Round 1 had low identity scores. Round 2 improved identity but lost motion.",
    "vqa_failure_analysis": [
      {{
        "vqa_question": "On a scale of 0-100, how clearly is the girl's red hair visible?",
        "score": 20,
        "category": "Reference Fidelity",
        "identified_flaw": "Hair Color Mismatch. The low score indicates the red hair from the reference is missing or wrong.",
        "refinement_action": "Explicitly add 'red hair' to the text prompt to enforce the reference color."
      }},
      {{
        "vqa_question": "On a scale of 0-100, how smooth is the camera movement without shaking?",
        "score": 45,
        "category": "Video Quality",
        "identified_flaw": "Camera Jitter/Instability.",
        "refinement_action": "Add 'smooth camera' and 'stabilized' to the prompt."
      }}
    ]
  }},
  "refinement_strategy": "The primary issues are the missing hair color (Reference) and camera shake (Quality). I will add specific visual descriptors for the hair and stability keywords.",
  "refined_prompt": "A cinematic video of a girl with vibrant red hair. She is walking forward. Smooth camera, stabilized, high quality."
}}
```

All the keys in this json example must be matched exactly in your response.

## Your input

---
[PROMPT REFINEMENT HISTORY]
{history}
[END PROMPT REFINEMENT HISTORY]
---

---
[ORIGINAL USER PROMPT]
{original_prompt}
[END ORIGINAL USER PROMPT]
---

---
[REFERENCE IMAGES]
{reference_images}
[END REFERENCE IMAGES]
---

---
[CURRENT ITERATION PROMPT]
{cur_iter_prompt}
[END CURRENT ITERATION PROMPT]
---

---
[CURRENT VIDEO]
{cur_video}
[END CURRENT VIDEO]
---

---
[QA PAIRS]
{qa_pairs}
[END QA PAIRS]
---

Please provide your analysis:

\end{lstlisting}
\end{agentbox}




\end{document}